\documentclass[conference]{IEEEtran}
\IEEEoverridecommandlockouts
\usepackage{amsmath,amssymb,amsfonts}
\usepackage{algorithmic}
\usepackage{graphicx}
\usepackage{textcomp}
\usepackage{xcolor}

\usepackage{microtype}
\usepackage{graphicx}
\usepackage{subfigure}

\usepackage{times}
\usepackage{epsfig}
\usepackage{graphicx}
\usepackage{amsmath}
\usepackage{amssymb}
\usepackage{graphicx}
\usepackage{times}
\usepackage{amsmath}
\usepackage[english]{babel}
\usepackage{graphicx}
\usepackage{epsfig}
\usepackage{epstopdf}
\usepackage{amsmath}
\usepackage[english]{babel}
\usepackage{graphicx}
\usepackage{graphicx}
\usepackage{epsfig}
\usepackage{epstopdf}
\usepackage{amssymb}
\usepackage{times}
\usepackage[small]{caption}
\usepackage{graphicx}
\usepackage{amsmath}
\usepackage{algorithm}
\usepackage{algorithmic}
\usepackage{amsthm}

\def\BibTeX{{\rm B\kern-.05em{\sc i\kern-.025em b}\kern-.08em
    T\kern-.1667em\lower.7ex\hbox{E}\kern-.125emX}}
\begin{document}

\title{Asynch-SGBDT:  Train a Stochastic Gradient Boosting Decision Tree in an Asynchronous Parallel Manner}
 
\author{\IEEEauthorblockN{Daning Cheng \IEEEauthorrefmark{1}\IEEEauthorrefmark{2},
		Fen Xia\IEEEauthorrefmark{3}, 
		Shigang Li\IEEEauthorrefmark{1} and
		Yunquan Zhang\IEEEauthorrefmark{1}}
	\IEEEauthorblockA{\IEEEauthorrefmark{1}SKL of Computer Architecture, Institute of Computing Technology, CAS, China\\
		Email:  \{chengdaning, lishigang, zyq\}@ict.ac.cn}
	\IEEEauthorblockA{\IEEEauthorrefmark{2}University of Chinese Academy of Sciences
	}
	\IEEEauthorblockA{\IEEEauthorrefmark{3}WiseUranium corp.\\
		{\{ Xiafen \}@ebrain.ai}}
}
\maketitle

\begin{abstract}
Gradient Boosting Decision Tree (GBDT) is an effective yet costly machine learning model. Current parallel GBDT algorithms generally follow a synchronous parallel design. Since the processing time for different nodes varies in practice, synchronisation in a parallel computing environment needs considerable time. In this paper, we propose an asynchronous parallel GBDT algorithm named as asynch-SGBDT. Our theoretical and experimental results indicate that compared with the serial GBDT training process, when the datasets are high-dimensional sparse datasets, asynch-SGBDT does not slow down convergence speed on the epoch. Asynch-SGBDT achieves 14x to 22x speedup when it uses 32 workers; LightGBM, as the benchmark, only achieves 5x to 7x speedup using 32 machines; Dimboost, as another benchmark, only achieves 4x to 5x speedup using 32 workers. All of theory and experimental results show that asynch-SGBDT is state-of-the-art parallel GBDT algorithm.
\end{abstract}

\begin{IEEEkeywords}
GBDT, asynchronous parallel, SGD, parameter server
\end{IEEEkeywords}

 \section{Introduction}
 
 Gradient Boosting Decision Tree (GBDT) is an effective yet costly nonlinear machine learning model. The demand for performance with efficient training process drives high-performance GBDT to be one of the most popular research areas in machine learning. Current methods in gaining computation efficiency rely heavily on parallel computing technology. However, the requirement for synchronisation creates a bottle-neck for further improvement in computation speed. In this paper, we propose an asynchronous parallel GBDT method that relaxes this requirement. Our asynchronous parallel stochastic GBDT (asynch-SGBDT) method works on the parameter server framework: The server receives trees from workers, and different workers build different trees in an asynchronous parallel manner. Our theoretical and experimental results indicate that when the datasets are high-dimensional sparse datasets, asynch-SGBDT exerts no impact on convergence speed on the epoch.  Our experimental results also indicate that asynch-SGBDT achieves a satisfied speedup. 
 
 Parallel computing is the most important technologies in gaining high computing performance. Current parallel GBDT algorithms mainly use fork-join parallel implementation, a kind of synchronous parallel manner, to accelerate the GBDT, like xgboost, LightGBM, and TencentBoost(DimBoost). 
 
 However, in practice, the synchronous parallel cannot reach satisfied scalability: LightGBM, the state-of-the-art parallel GBDT framework, achieves 5x to 7x speedup when it uses 32 machines. In a cluster, it is unlikely that all nodes in a system share the same computation speed. Faster nodes have to be blocked before all nodes reach the barrier. Thus, synchronisation needs a considerable amount of time. This problem exerts more influence on synchronous parallel GBDT, as the synchronous operation is contained in every GBDT iteration step.

 Asynchronous parallel methods are the most important parallel methods to break the shortage of synchronous. Research in this area seems to have ignored an important question: Is synchronization necessary for GBDT training, and if a GBDT can be trained in an asynchronous parallel manner? Are asynchronous parallel methods effective in GBDT training process?
 
 
 We develop theoretical analyses via connecting asynch-SGBDT training process with stochastic optimization problem: we adopt functional space optimization into high-dimensional parameter space optimization following \cite{Sun2014A}, and we introduce a random variable into the objective function through sampling operation. Thus, asynchronous parallel stochastic gradient descent method can be used to train GBDT.

Theoretical analyses show that high diversity samples of the datasets, small sampling rate,  small step length and a large number of leaves for the GBDT settings can lead to high scalability for asynch-SGBDT. We call such conditions as asynch-SGBDT requirements. 

Our analyses and experiments suggest that compared with the serial GBDT training process, when the datasets are high-dimensional sparse datasets, asynch-SGBDT does not slow down convergence speed on the epoch.  The widely used datasets in the big data industry, which are high-dimensional and sparse, are likely to meet the asynch-SGBDT requirements.

What is more, asynch-SGDBT reaches 14x-20x speedup when it uses 32 workers on real-sim and E2006-log1p dataset. As the benchmark, LightGBM, a state-of-art parallel GBDT framework, only achieves 5x-7x speedup when it uses 32 machines.

Our main contributions are the following: (1) We propose asynch-SGBDT, an asynchronous parallel method that can be used to train the GBDT model more efficiently. (2) We provide theoretical justification for asynch-SGBDT through an analysis of the relationships among convergence speed, number of workers and algorithm scalability. (3) Our approach achieves higher efficiency and accuracy than state-of-the-art on our tested datasets, which are high dimensional and sparse.

 \section{Related Works and the Problem of Current Algorithm}
 The standard GBDT iteration step contains two sub-steps: producing the target sub-step and building the tree sub-step. Besides standard algorithm steps \cite{Friedman2001Greedy}, many studies have offered alternative and improved operations in the above two sub-steps. Sampling strategies and novel pruning strategies  \cite{Kalal2008Weighted} \cite{Appel2013Quickly} and parallelization technology on the building tree process are the main methods to accelerate the GBDT training process  \cite{Chen2016XGBoost}  \cite{NIPS2017_6907}.

 Sampling is an alternative operation in producing the target sub-step. A number of studies have described different sampling strategies in the GBDT training process   \cite{NIPS2017_6907} \cite{Gupta2004Boosted} \cite{Kalal2008Weighted}.  GBDT based on sampling  \cite{Friedman2002Stochastic}  \cite{Ye2009Stochastic} also is named as stochastic GBDT. These stochastic GBDT algorithms focus on how to build sampling subdatasets with good performance that represent the whole dataset well.

 \textbf{Problem: How to train a GBDT in an asynchronous parallel manner} Some researchers have proposed a stochastic gradient boosting tree \cite{Friedman2002Stochastic} and a parallel stochastic gradient boosting tree \cite{Ye2009Stochastic}. However, these methods use fork-join parallel methods. Asynchronous parallel SGD \cite{Langford2009Slow}   \cite{Niu2011HOGWILD} and its implement frame,  parameter server \cite{li2014scaling}, are the most widely used asynchronous parallel framework. Although J Jiang et al. \cite{Jiang2017TencentBoost} tried to adapt the GBDT on a parameter server (i.e., TencentBoost), TencentBoost still uses the fork-join parallel method. The parallel part only exists in the sub-step of building the tree. The fork-join method fails to make full use of the performance of parameter servers. Currently, few works focus on the question of whether it is possible to use asynchronous parallel methods to train the GBDT model.
 
 In this paper, we examine whether it is possible to use asynchronous parallel methods to train GBDT model, how to train GBDT model in an asynchronous parallel manner and find the conditions under which GBDT is trained in an asynchronous parallel manner effectively.
 
 \section{Background}

 \subsection{Boosting and Its Theory}
 
 \subsubsection{Problem Setting, Serial Training Method}
 Given a set of training data $\left\lbrace( x_{i},y_{i})\right\rbrace ^{\sum_{j=1}^{N}m_j}$,
 where the feature $x_{i}\in \mathcal{X}$, the label $y_{i}=\left\lbrace 1,0\right\rbrace $, and $m_j$ represents the frequency of $(x_j,y_j)$ in the dataset,  the variables $(x_j,y_j)$ are  different from each other.
 
 The goal for boosting is to find a predictor function for the additive classifier (i.e., GBDT forest), 
 $F=F(x)\in R$, by minimizing the total loss over the training dataset
 
 \begin{align}
 \label{boostinggoal}
 min_{\mathbf{F}\in \mathbb{R}^N}L(\mathbf{F})\mbox{ },   L(F):=\sum_{i=1}^{N}m_i\ell(y_{i},F(x_{i})).
 \end{align}
 
 
 where $F_{i}$ is shorthand for $F(x_{i})$ and $\ell'_i$ is shorthand for $\partial_{F_i}\ell(y_i,F_i)$ in the following part of the
 article. Theoretically, $\ell(\cdot,\cdot)$ can be an arbitrary and convex surrogate function. In Friedman et al.'s work and based on the requirement in machine learning,  $\ell(\cdot,\cdot)$ adopts the logistic loss:
 
 \begin{align*}
 &\ell(y_i,F_i) = y_ilog(\frac{1}{p})+(1-y_i)log(\frac{1}{1-p}),p=\frac{e^{F_i}}{e^{F_i}+e^{-F_i}}
 \end{align*}
 This additive function $F(x)$ (i.e., GBDT forest) produces
 the vector $\mathbf{F}=[F_{1},F_{2},...,F_{N}]$ to minimize $L(\mathbf{F})$. This process is an optimization on the $R^{N}$ parameter space.
 
 We define $\mathbf{F^*}$ and $\mathbf{L^*}$  as follows.
 
 \begin{equation}
 \mathbf{F^*}=\mathop{argmin}L(\mathbf{F}),\mathbf{L^*}=\mathop{min}L(\mathbf{F})
 \end{equation}

 The whole process of the iteration step in traditional GBDT training is divided into two sub-steps:  1. Producing a target for the building tree. For example, the target is the gradient vector of the dataset on a certain loss \cite{Sun2014A}, i.e. the vector as follow:
 \begin{equation}
 \label{gradient_dataset}
 \mathbf{G}=[m_1\ell_1',m_2\ell_2',...m_N\ell_N']
 \end{equation}
 2. Building  a  tree whose output is close to $\mathbf{G}$ (the prediction for $x_i$ is close to $\ell_i'$ in this tree). 
 
 How to build a tree (i.e., the detail of building a tree sub-step) is not the topic in this paper.

 \subsubsection{Parallel GBDT Training Algorithm}
 Current GBDT algorithm uses fork-join parallel implementation to parallelize the GBDT training process. Only the parallel parts stay in the building tree sub-step process in the current GBDT framework and algorithm, such as xgboost \cite{Chen2016XGBoost} and lightGBM \cite{NIPS2017_6907}. 
 
 \subsection{GBDT trees}
 Decision trees are an ideal base learner for data mining applications of boosting (Section 10.7 in the book \cite{Hastie2015The}). Thus, this subsection, we would present the necessary theories of the decision tree.

 The goal of building a decision tree is to build the tree whose leaves contain the same samples. For the GBDT case, the above statement is that, for the decision tree, the prediction for $x_i$ is $\ell_i'$ in Eq. \ref{gradient_dataset}. Without regard to generalisation, a well-grown decision tree often reaches above goal.

Above statements are equal to the statement that each leaf reach the minimum RMSE or misclassification rate. 

The book\cite{Hastie2015The} presents more detail information about the decision tree. 

 \subsection{SGD and Its Parallel Engineering Methods}
 
 SGD is used to solve the  following problem \cite{Duchi2016Intro}
 
 \begin{align}
 \label{sgd_define}
 &minimize_{w\in X}f(w)\mbox{ },f(w):=E\left[Function(w;\mathbf{\mathfrak{\textrm{\ensuremath{\Theta}}}})\right]\\
 &E\left[Function(w;\mathbf{\Theta})\right]=\int_{\Xi}Function(w;\theta)dP(\theta)
 \end{align}

 where a random variable $\mathbf{\Theta}$ has a  probability distribution function (PDF) $dP(\theta)$. We use the frequency instead of $dP(\xi)$, which means

 \begin{equation}
 f(w)\approxeq\frac{1}{n}\sum_{1}^{n}Function(w,\theta_{i})
 \label{sgd_pri}
 \end{equation}

 The algorithm of the delayed SGD (i.e., an asynchronous parallel SGD algorithm) \cite{Langford2009Slow} \cite{Niu2011HOGWILD} is described as the most important parallel SGD. The implements of delayed SGD, (parameter server)  include ps-lite in MXNET \cite{Chen2015MXNet}, TensorFlow \cite{Abadi2016TensorFlow}, and petuum \cite{Xing2015Petuum}.
 
 \textbf{Random variable} In a traditional machine learning problem, such as training a support vector machine (SVM) model, an observed value  $\theta_{i} $ of random variable $\mathbf{\Theta}$ is a sample in the dataset.

 \subsubsection{Serial SGD}
 To solve the above problem, SGD, as shown in algorithm \ref{serial_sgd}, is the most widely used method because of its small memory requirement and fast convergence speed.
 \begin{algorithm}
 	\caption{stochastic gradient descent. }
 	\label{serial_sgd}
 	\begin{algorithmic}
 		\STATE {\bfseries Input:}dataset $\{ \theta_1,\theta_2...\theta_n \}$,Learning Rate $v$
 		\STATE {\bfseries Output:}$w_n$
 		\FOR{ $t=1,2...n$ }
 		\STATE $w_{t+1}=w_t-v\partial_wFunction(w_{t},\theta_{t})$
 		\ENDFOR
 	\end{algorithmic}
 \end{algorithm}

 \subsubsection{Asynchronous Parallel SGD and Parameter Server Framework}

 The base of parameter server is delayed SGD.
 
 Delayed SGD gains high performance via the overlapping of the communication time and computation time. 
 \begin{algorithm}
 	\caption{Asynchronous parallel stochastic gradient descent (i.e., delayed SGD). }
 	\label{sgd_delay}
 	\begin{algorithmic}
 		\STATE {\bfseries Input:}dataset $\{ \theta_1,\theta_2...\theta_n \}$,Learning Rate $v$
 		\STATE {\bfseries Output:}$w_n$
 		\FOR{ $t=1,2...n$ }
 		\STATE $w_{t+1}=w_t-v\partial_wFunction(w_{k(t)},\theta_{k(t)})$
 		\ENDFOR
 	\end{algorithmic}
 \end{algorithm}
 
 \subsection{Sampling and Stochastic GBDT}
 Many studies have also proposed the stochastic GBDT algorithm which uses sampling strategies. GBDT algorithms based on sampling datasets are also named as stochastic GBDT algorithms\cite{Friedman2002Stochastic}  \cite{Ye2009Stochastic}. However, the above studies fail to involve asynchronous parallel methods. They use sampling strategies to reduce the burden of building a tree to improve the accuracy of the GBDT output \cite{NIPS2017_6907} \cite{Gupta2004Boosted} \cite{Kalal2008Weighted}.    These works focus on how to make sample subdatasets that share the same characteristics with the full dataset.

 \section{Main Idea:The Equivalent of  Stochastic GBDT and the Stochastic Optimization}
 
 In this section, we will prove the following corollary.
 
 \textbf{Corollary 1} Training stochastic GBDT and stochastic optimization are the equivalence problem. 
 
 \begin{proof}
     In stochastic GBDT, each iteration step is described as follows: 
 1. Sampling the data:  Each sample in the dataset corresponds to a Bernoulli distribution. In each iteration step, selecting this sample depends on its Bernoulli distribution. Traditional stochastic GBDTs view sub-datasets and full datasets share the same characteristics. 2. Build target: In the gradient step, stochastic GBDT calculates the gradient of a subdataset on a certain loss.
 3. Building the tree based on the target.
 
 However, we treat the sampling process as the method which introduces a random variable into the original objective function. Then, the goal for stochastic GBDT is to find an additive function $\mathbf{F}$ by minimising the mathematical expectation of the total loss over the training dataset. 

 	\begin{align}
 	\label{L_random}
 	&minimize_{\mathbf{F}\in \mathbb{R}^N}f(\mathbf{F})\mbox{ } ,f(F):=E\left[L_{random}(\mathbf{F};\mathbf{\mathfrak{\textrm{\ensuremath{\mathbf{Q}}}}})\right] \notag\\
 	&{L_{random}}(\mathbf{F};\mathbf{Q}):=\sum_{i=1}^{N}(\sum_{j=1}^{m_j}\frac{Q_{i,j}}{R_{i,j}})\ell(y_{i},F_{i})
 	\end{align}

 	where $\mathbf{Q}=(Q_{1,1}..Q_{1,m_1},Q_{2,1},...,Q_{2,m_2},Q_{N,1},...,Q_{N,m_N})$ and $Q_{i,j}$ is a random
 	variable that satisfies the Bernoulli distribution: $P(Q_{i,j}=1)=R_{i,j}$, $P(Q_{i,j}=0)=1-R_{i,j}$.
 	
 	In every sampling process,  the sampling operation in the stochastic GBDT would produce an observed value vector corresponding to $\mathbf{Q}$. Based on the observed value vector, the sampling operation in stochastic GBDT produces the sampling subdataset.
 	
 	Combining the convex characteristics of $L$ and $L_{random}$, the  following expressions are true.
 
 	\begin{align}
 	\label{core_equ}
 	\mathop{min} E[L_{random}] = \mathop{min} L \\
 	\label{core_equ1}
 	\mathop{argmin} E[L_{random}] = \mathop{argmin} L
 	\end{align}

 	Now, our optimisation objective function is changed from $L(\mathbf{F})$ to $L_{random}(\mathbf{F};\mathbf{Q})$, which means training stochastic GBDT and stochastic optimisation are the equivalence problem. 
 \end{proof}

 \textbf{Using SGD to Solve Stochastic GBDT}      Corollary 1 shows that it is possible to use high-performance stochastic optimisation algorithm, like asynchronous parallel SGD to train stochastic GBDT.  
 
\begin{figure}[th]
 	
 	\centering
 	\includegraphics[width=0.5\textwidth]{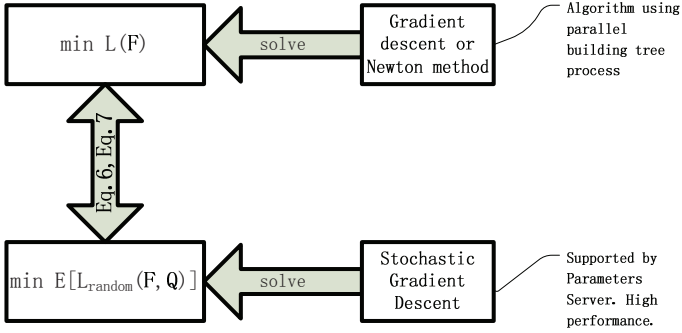}
 	\caption{Using SGD to train the stochastic GBDT\textbf{:} The minimums of $L$ and $E[L_{random}]$ are the same. Solving the minimum of $E[L_{random}]$ is a high-performance process.}	
 	\label{obj_change}	
 \end{figure}

 Eq. \ref{core_equ} shows that   $\mathbf{Q}$ is a random variable vector and $\mathbf{F}$ is a variable.  The definition of $L_{random}$ in Eq. \ref{L_random} is the same as  the expression of $Function(w;\mathbf{\Theta})$ in Eq.\ref{sgd_define}.  The above facts suggest that it is possible to use well parallelised stochastic optimisation, such as the asynchronous parallel SGD algorithm, to find the minimum of $L_{random}$. Above process is described by Figure \ref{obj_change}.     
 
 \textbf{Random Variable} In traditional machine learning problems, such as training an SVM model, an observed value of a random variable is a sample in the dataset. In asynch-SGBDT, an observed value of a random variable is the sampling result of the sampling subdataset. 
 
 \section{Asynch-SGBDT and its Analyses}

 \subsection{Asynch-SGBDT}
 
 \begin{figure}[th]
 	\centering
 	\includegraphics[width=0.5\textwidth]{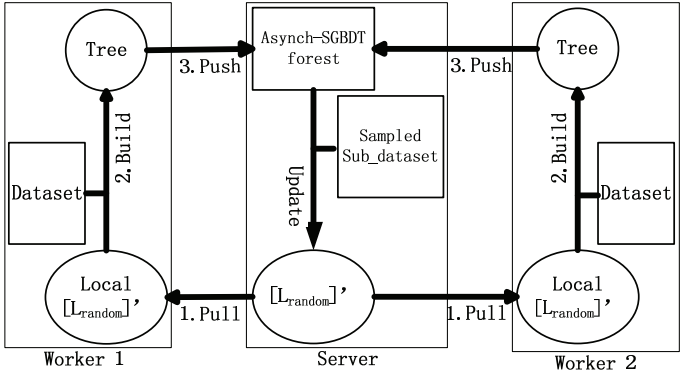}
 	\caption{Asynch-SGBDT on the parameter server\textbf{:} worker 1 and worker 2 are asynchronously parallel and independently work. The server updates $L'_{random}$ at once when it receives a tree from any worker.}
 	\label{asynch-SGbdt_ps}
 	
 \end{figure}
 Asynch-SGBDT uses asynchronous parallel SGD to train the model. Asynch-SGBDT is described by algorithm \ref{boostingps}. The work model of asynch-SGBDT is illustrated in Figure \ref{asynch-SGbdt_ps}. The sequence diagrams of asynch-SGBDT are shown in the last sub-figure in Figure \ref{timemode}. In Figure \ref{timemode}, we also show the sequence diagrams of the different GBDT training methods.

 In algorithm \ref{boostingps}, $L_{random}'$ is  shorthand for the SGD of $L_{random}(\mathbf{F};\mathbf{Q})$ and  $L_{random}'$ is calculated as follows
 
 \begin{equation}
 {L_{random}'}:=[m'_1\ell_1',m'_2\ell_2',...,m'_N\ell_N']
 \label{Lrangdom'}
 \end{equation} 
 
 where  $m'_i=\sum_{j=1}^{m_i}\frac{Q_{i,j}}{R_{i,j}}$. In building the tree sub-step, the algorithm still builds  the  tree whose  prediction for $x_i$ is close to $\ell_i'$.

 \textbf{Asynchronous parallel} Asynchronous parallel means that different workers are blind to each other.  These workers work independently. The pull, build, and push operations of a worker must be ordered and serialised but, for different workers, these operations in different workers are completely out of order and parallel. During the time that a worker is building the $i$th tree, $F^t(x)$ and $L'^t_{random}$  in the server would be updated several times by other workers.  If the number of workers is large enough, the building tree sub-step would be hidden. Additionally, compared with the original datasets, the size of the sampling subdataset is relatively small, which would reduce the burden of  the building tree sub-step.
 
 
 \begin{algorithm}[!th]
 	\caption{  Asynch-SGBDT }
 	\label{boostingps}
 	\begin{algorithmic}
 		\STATE {\bfseries Input:} {$\left\lbrace x_{i},y_{i}\right\rbrace ^{N}$: The training
 			set; $v$: The step length;}
 		\STATE {\bfseries Output:}{the Additive Tree Model ${F}=F(x)$, i.e. asynch-SGBDT forest.}

 		\STATE {\textbf {For Server:}}
 		
 		Produce the tree whose output is $\frac{1}{\sum_{i=0}^{N}m_{i}}\sum_{i=0}^{N}{m_{i}y_i}$.
 		
 		Calculate  $L'^{0}_{random}$ and Maintain $L'^{0}_{random}$.
 		
 		\FOR{$j=1...forever$}
 		
 		\STATE 1.Recv a $Tree_{k(j)}$ from any worker, this tree is built based on $L'^{k(j)}_{random}$.
 		
 		\STATE 2.Add $Tree_{k(j)}$ times $v$ to whole asynch-SGBDT forest.\; (${F}^j(x)= {F}^{j-1}(x) + vTree_{k(j)}$)
 		
 		\STATE 3.Generate an observed value vector of $\mathbf{Q}$ and produce sampling sub-dataset.
 		
 		\STATE 4.Calculate current GBDT forest's  $L'^{j}_{random}$ based on  sampling sub-dataset.
 		
 		\STATE 5.Remove $L'^{j-1}_{random}$ and Maintain $L'^{j}_{random}$ ($L'^{j}_{random}$ can be pulled by workers.).
 		\ENDFOR	
 		\STATE {\textbf {For Worker:}}
 		
 		\FOR{$t=1$ to forever}
 		\STATE  1.Pull the $L'^{t}_{random}$ from Server ($L'^{t}_{random}$ is current $L'_{random}$ the Server holds.).
 		\STATE 2.Build $Tree_{t}$ based on $L'^t_{random}$.
 		\STATE 3.Send $Tree_{t}$ to Server.
 		
 		\ENDFOR

 	\end{algorithmic}
 \end{algorithm}
 \subsection{Convergent analysis of Asynch-SGBDT}

 \subsubsection{Special Case}
 To make our presentation clear, we show a result of a special case which would show how different variable intuitively without paying too much cost. General case and conclusion will offer in the next subsection.

 Our analysis in this subsection should be based on the assumption that each decision tree in GBDT is a well-grown tree. This assumption is reasonable because most of the GBDT building tree sub-step tries its best to fit the gradient in the gradient descent method or the vector of the gradient times the inverse Hessian matrix in Newton descent methods\cite{Chen2016XGBoost}, like section 3.2. 
 
 We introduce additional math definitions as follows: $Lip$ is the Lipschitz constant for $L(.)$. $L(.)$ is a strong convex function with modulus $c$. For $\mathbf{F} \in \mathbb{R}^N$ and every observed value vector corresponding to $\mathbf{Q}$, $\|L'_{random}\|<M$. The random variable vector $\mathbf{Q'}$ is defined as
 \begin{align*}
 \mathbf{Q'}=[Q'_1,Q'_2,...,Q'_N]
 \end{align*} 
 where the random variable  $Q'_i = (Q_{i,1} \lor Q_{i,2} \lor...\lor Q_{i,m_i})$. $\Omega = \mathop{max} \sum_{i=1}^{N}Q'_i$, which represents the maximum number of different samples in one sampling process. This value is almost equal to the number of different samples in the dataset because it is possible that all samples are sampled in a sampling subdataset. $\Delta = \mathop{max} P(Q'_i = 1)$, which is the maximum probability that a type of sample is sampled. $\rho$ represents the probability that two sampling subdatasets, whose intersection is nonempty, exist throughout the whole sampling process.
 
 \begin{figure}[tb]
 	\centering
 	\hspace{-0.8cm}\includegraphics[width=0.5\textwidth]{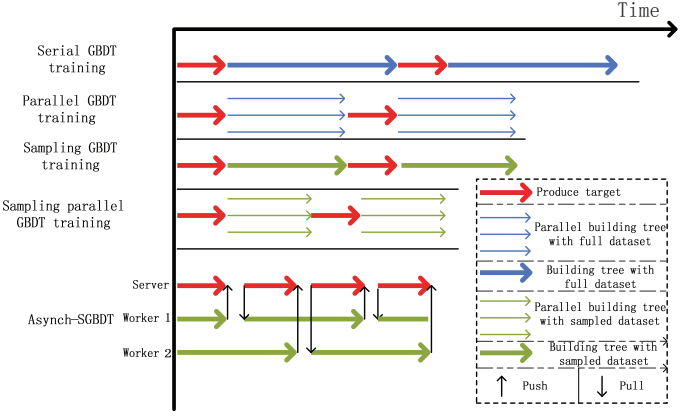}
 	\caption{Different GBDT Training Method Patterns}	
 	\label{timemode}
 \end{figure}
 We apply the proposition of asynchronous SGD  \cite{Niu2011HOGWILD} to Algorithm \ref{boostingps}.

 \textbf{Proposition 1.} Suppose in Algorithm \ref{boostingps} that
 
 $(1)$ $ \tau \geq j-k(j)$, $(2)$ $v$ is defined as
 
 \begin{equation}
 v=\frac{c\vartheta\epsilon}{2LipM^2\Omega(1+6\rho\tau+4\rho\tau^2\Omega\Delta^{1/2})}
 \end{equation}
 $(3)$  By defining $D_0:=\|\mathbf{F}_0-\mathbf{F}^*\|^2$, $\mathbf{F}_t$ is the $\mathbf{F}$ vector produced by $F^t(x)$.  $(4)$  For some $\epsilon >0$ and $\vartheta \in (0,1) $, $t$ is an integer satisfying
 
 \begin{equation}
 t\geq \frac{2LipM^2\Omega(1+6\rho\tau+6\rho\tau^2\Omega\Delta^{1/2}log(LD_0/\epsilon ))}{c^2\vartheta\epsilon}
 \end{equation}
 
 Then after $t$ updates in the server,
  
 \begin{equation*}
 E[L(\mathbf{F}_t)-L^*]<\epsilon.
 \end{equation*}

 \textbf{Analysis: Convergence Speed}    Proposition 1 shows that stochastic GBDT converges to a minimum of $L(\cdot)$. However, the influence of the sampling rate is unclear in serial stochastic GBDT: When $\tau = 0$, no elements in Proposition 1 would be changed with the change in sampling rate. $\Omega$ maybe be influenced by different sampling rates in Proposition 1. However, if the number of iterations is large, the impact would be small. Proposition 1 also shows that with the increase in the number of workers (i.e., the number of delays), a smaller step length should be chosen, and asynch-SGBDT needs more iteration steps to reach a satisfactory output.
 
 
 \textbf{Analysis: Scalability Upper Bound} It is easy to see that  the max number of worker algorithm satisfies following upper bound.

 \begin{equation}
 \#(worker)<\frac{T(Build Tree)}{T(Communicate+BuildTarget)}
 \label{max_speedup}
 \end{equation}

 where $\#(worker)$ is the max number of workers, $T(operation)$ is the time of operations. This inequation can be got from the condition that  computing time for each worker is fully overlapped from the computing time in the server.

 \textbf{Analysis: Scalability and Sensitivity} The sensitivity of the mathematical convergence speed to the change in the number of workers is another index that measures the scalability of the algorithm. If the mathematical convergence speed is insensitive to the number of workers (i.e., parallelism), the algorithm allows us to use more workers to accelerate the training process and gain a greater speedup.
 
 In the experiments, we notice that the sensibilities of the convergence speed to the change in the worker setting are different under different sampling rate settings and datasets. Our proposition shows that sensitivity is linked to $\rho$ and $\Delta$.  The sparsity of the observed value vector of $\mathbf{Q'}$  is positively correlated to the values of $\rho$ and $\Delta$ when sampling rates between different samples are almost the same or uniform. Therefore, it is possible to offer guidance and draw conclusions by analyzing the observed value vector of the sparsity of $\mathbf{Q'}$.

 \begin{figure*}[tb]
 	\centering
 	\subfigure[The original dataset contain samples: 10000 * $A_1$, 20000*$A_2$ 30000*$A_3$. The sample diversity of orighinal dataset is low]{
 		
 		\includegraphics[width=0.47\textwidth]{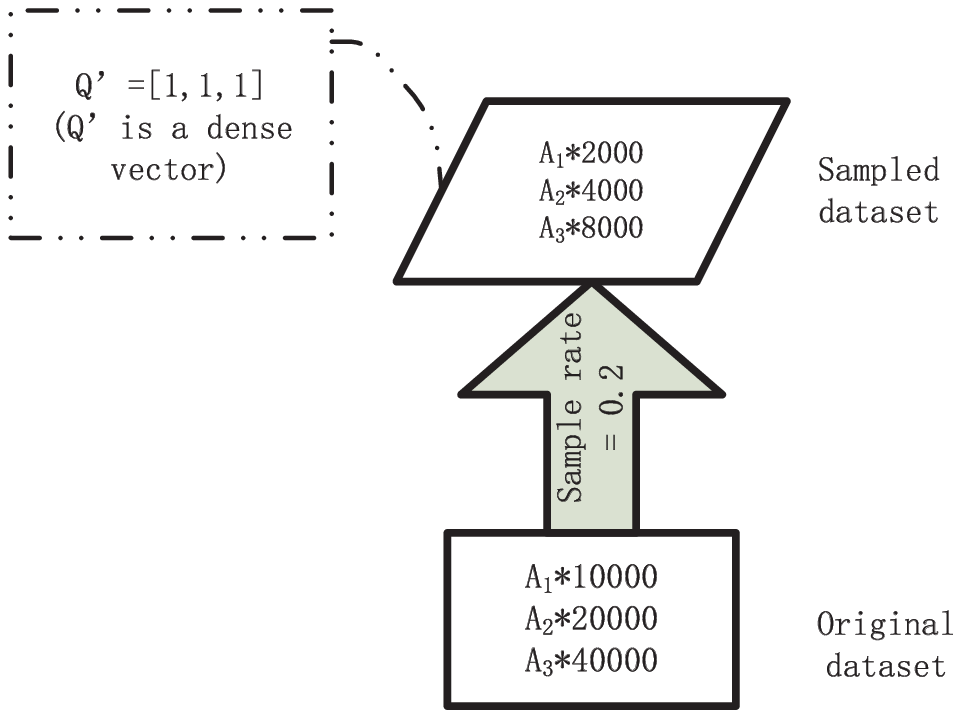}
 		
 	} 
 	\subfigure[The original dataset contain samples: $A_1$ ... $A_{14000}$ and each sample only appears once. The sample diversity of orighinal dataset is high ]{
 		\includegraphics[width=0.47\textwidth]{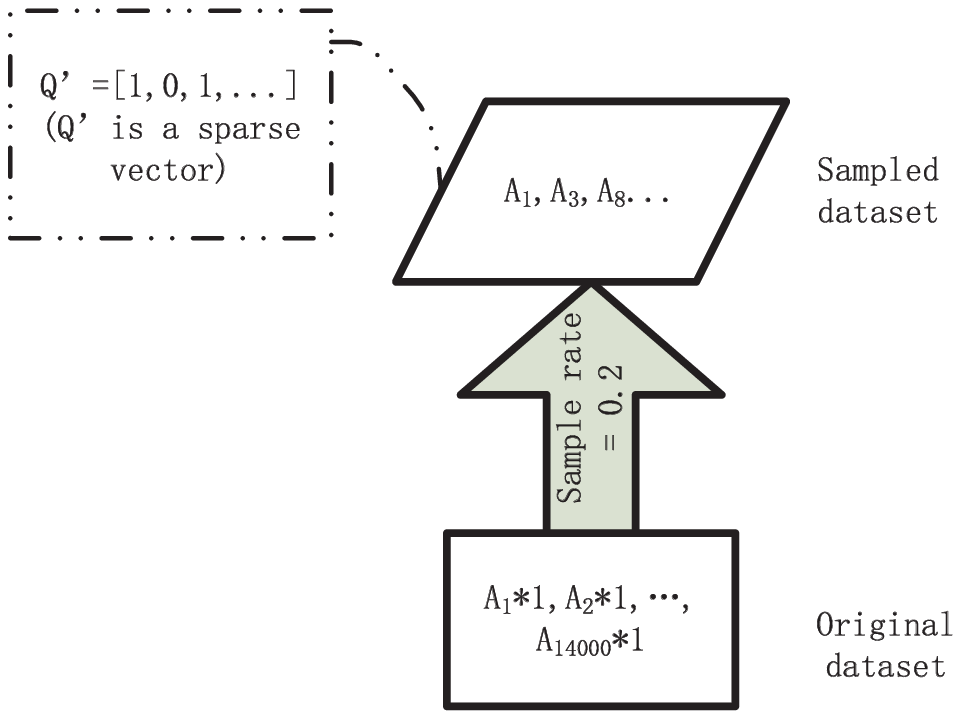}
 	}
 	\caption{The  sample diversity in dataset exert influence on the  the sparsity of $\mathbf{Q}$}	
 	\label{diversity}
 \end{figure*}

 The size of $\mathbf{Q'}$ represents the diversity of the samples in the dataset (i.e., the number of sample species). The number of non-zero elements in each sampling process represents the diversity of the samples in the sampling subdataset. If the diversity of the dataset is high, but the diversity of the sampling subdataset is low, the observed value vector of  $\mathbf{Q'}$ is likely to be a sparse vector in each sampling process, which means that the values of $\rho$ and $\Delta$ would be small. The algorithm would be insensitive to the number of workers. Reducing the sampling rates would help reduce the diversity of the sampling subdataset. Using a high-dimensional sparse dataset would contribute to increasing the diversity of the dataset. For the GBDT tree, different samples would be treated as the same sample if the number of leaves is too small. In this case, a small number of leaves in the GBDT tree would decrease the diversity of the dataset. For a low diversity dataset, even using the small sampling rate,  the observed value vector of  $\mathbf{Q'}$ is still a dense vector, just like the illustration of Figure \ref{diversity}. 
 Therefore, asynch-SGBDT is able to accelerate the high-dimensional sparse dataset with a relatively small sampling rate. The high-dimensional sparse dataset is the most frequently used dataset in the era of big data.

  \subsubsection{General case}
In our analysis in  Appendix part, we show the detail of the analysis of asynch-SGBDT convergence process.

In our analysis, we can gain the following conclusion for any asynch-SGBDT setting:

\textbf{General Conclusion}  We summarize the general conclusions as follows: 

1. Small sampling rate would decrease the convergence speed. However, small sampling rate help decrease the sensitivity of the algorithm, which would increase the scalability.

2. Under an asynchronous parallel situation and using a fixed sampling rate, the more workers we use, the smaller the step length that should be chosen, and the more steps the algorithm would run. 

3. Under an asynchronous parallel situation and using a fixed number of workers, the larger the sampling rate is, the more sensitive the algorithm. 

4. Under an asynchronous parallel situation and using a fixed number of workers, the smaller the sampling rate is, the larger the step length that should be chosen, and the fewer steps the algorithm would run. 

5. Asynch-SGBDT is apt to accelerate the GBDT  on a high-dimensional sparse dataset. 

6. Asynch-SGBDT is apt to accelerate a GBDT whose trees contain massive leaves. 

We also can draw the requirements of high scalability for asynch-SGBDT: high dataset diversity, small sampling rate and large learning rate, large GBDT leaves number setting. Those requirements are named as asynch-SGBDT requirements in this paper. Usually, normal GBDT settings on large scale dataset satisfy asynch-SGBDT requirements.

Besides the above general conclusions, we also can draw the following counter-intuitive conclusions: 1. The sampling process is necessary for asynchronous parallel; even we have enough computing resource.
2. Only gradient step can use asynchronous parallel manner. Thus, xgboost cannot be modified into asynch-parallel manner. 
 
 \subsection{Compatibility for the Parameter Server}

 Figure \ref{timemode} shows different training iteration steps of the different GBDT algorithms and framework. The traditional GBDT training algorithm is a serial process. Just as we mentioned in the introduction section, the iteration step contains two sub-steps: Producing the target and building the tree. Based on the traditional training algorithm, current GBDT frameworks use parallel or sampling technology in the building tree sub-step.  A number of frameworks only use parallel, distribution methods, such as xgboost \cite{Chen2016XGBoost} and TencentBoost \cite{Jiang2017TencentBoost}.  Many GBDT algorithms use sampling technology to reduce the burden of building tree sub-steps, such as stochastic gradient boosting \cite{Friedman2002Stochastic}. Using sampling and parallel technologies together is the most popular method to accelerate the building tree sub-step, such as LightGBM \cite{NIPS2017_6907} and stochastic GBDTs \cite{Ye2009Stochastic}. 
 
 However,  the order of producing the target sub-step, building tree sub-step, and all iteration steps is a rigorous serial process in the above framework. Whole iterations can only be parallelized via a fork-join parallel model.  The fork-join style fails to make full use of the parameter server's performance.  The parameter server is the basis of the machine learning and artificial intelligence (AI) industry. 
 
 Asynch-SGBDT breaks the above serial limitations. The base of asynch-SGBDT is the same as the base of the parameter server (i.e., the asynchronous parallel stochastic gradient algorithm). Different building tree sub-steps, produced target sub-steps and communication overheads are overlapped in asynch-SGBDT. Given the parallel mode, asynch-SGBDT is innovative in the GBDT training process. Asynch-SGBDT provides good compatibility with the parameter server.

 \section{Experiment }
 To prove our statements that asynch-SGBDT is state-of-the-art parallel GBDT algorithm, we conducted two experiments: validity experiment and efficiency experiments.
 
 \textbf{Validity experiments:} To prove the algorithm validity and correspond with the asynch-SGBDT theoretical analyses, we conducted comparison experiments. In the experiments, we used asynch-SGBDT to deal with the two classification problems. In our comparison experiments, we chose the dataset and GBDT tree settings that did and did not meet the asynch-SGBDT requirements. The ideal experimental result is that the former experiments exhibit high scalability (i.e., low sensitivity), and the latter experiments exhibit low scalability (i.e., high sensitivity).
 
 \textbf{Efficiency experiments:} To prove the asynchronous parallel method is better than the current parallel GBDT method, we compared the speed up ratio of asynch-SGBDT with feature parallel. Feature parallel is the main parallel method in LightGBM \cite{NIPS2017_6907}. 
 
 
 \subsection{Datasets}
We chose three datasets: the real-sim dataset and the Higgs dataset, which were selected from the SVM library (LIBSVM) repository as our experimental dataset. Real-sim and E2006-log1p datasets are large datasets, where sample vectors are high-dimensional and sparse. The Higgs dataset is a large dataset, where sample vectors are low-dimensional and dense. 

We use real-sim dataset and Higgs dataset in validity experiment.  
The Higgs dataset does not meet asynch-SGBDT requirements 
Thus, the Higgs experiments are treated as benchmark experiments in validity experiment. We use real-sim dataset and E2006-log1p dataset in efficiency experiment. 

We used 100000 samples from the whole dataset as the test dataset in the Higgs experiments. We used 16000 samples from the whole dataset as the test dataset in the real-sim experiments. We used 16,087 samples as the train dataset and     3,308 samples as the test data in the E2006-log1p experiments.
 
 %
 \subsection{Validity experiment}
 \subsubsection{Experiment Settings}
 
 In the real-sim experiments, we built 400 trees in total, and each tree had a maximum of 100 leaves. In the Higgs experiments, we built 1000 trees in total, and each tree had a maximum of 20 leaves. We randomly sampled 80\% of features in the experiments to build a tree at each building tree sub-step. The step length ($v$) in the experiments was fixed at 0.01. To gain clear experimental results, we set all sampling rates ($R_{i,j}$) to be the same. In our experiments, threads played the role of workers.
 
 \subsubsection{Analysis of Experimental Results}
 \begin{figure}[!tb]
 	\centering
 	\subfigure[sampling rate of 0.2 ]{
 		
 		\includegraphics[width=0.47\textwidth]{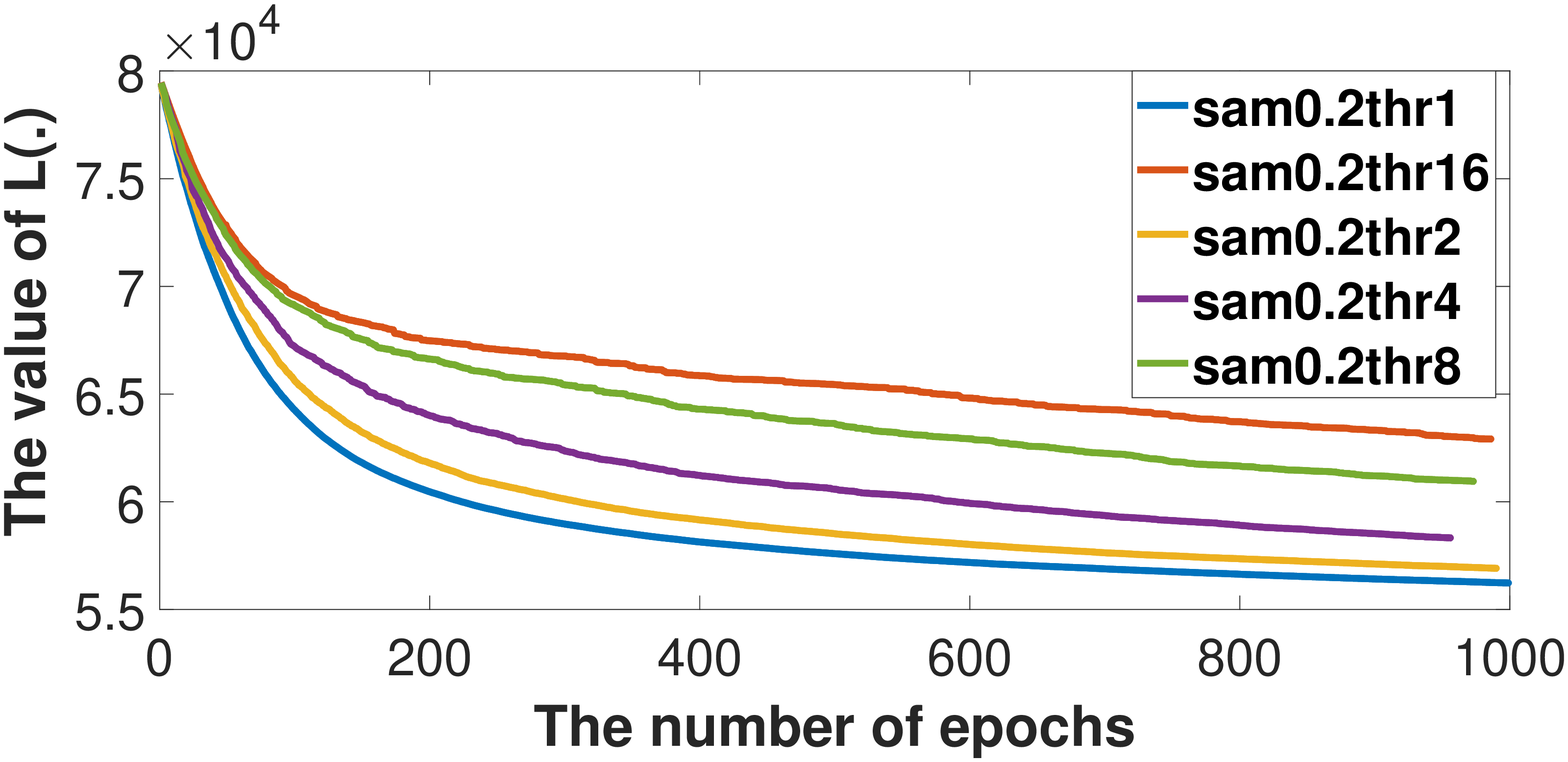}
 		
 	} 
 	\subfigure[sampling rate 0.4 ]{
 		\includegraphics[width=0.47\textwidth]{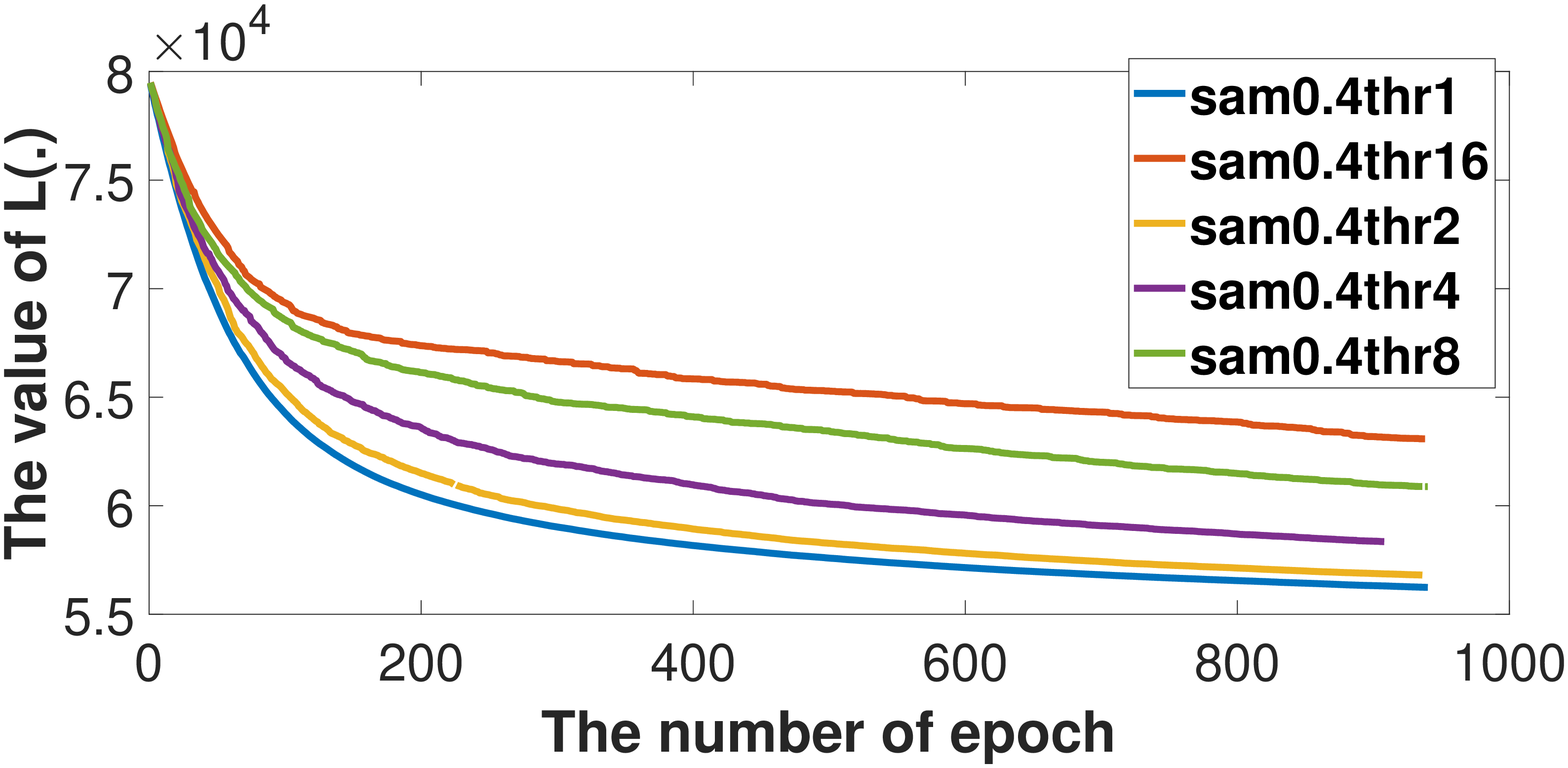}
 	}
 	\subfigure[sampling rate 0.6 ]{
 		\includegraphics[width=0.47\textwidth]{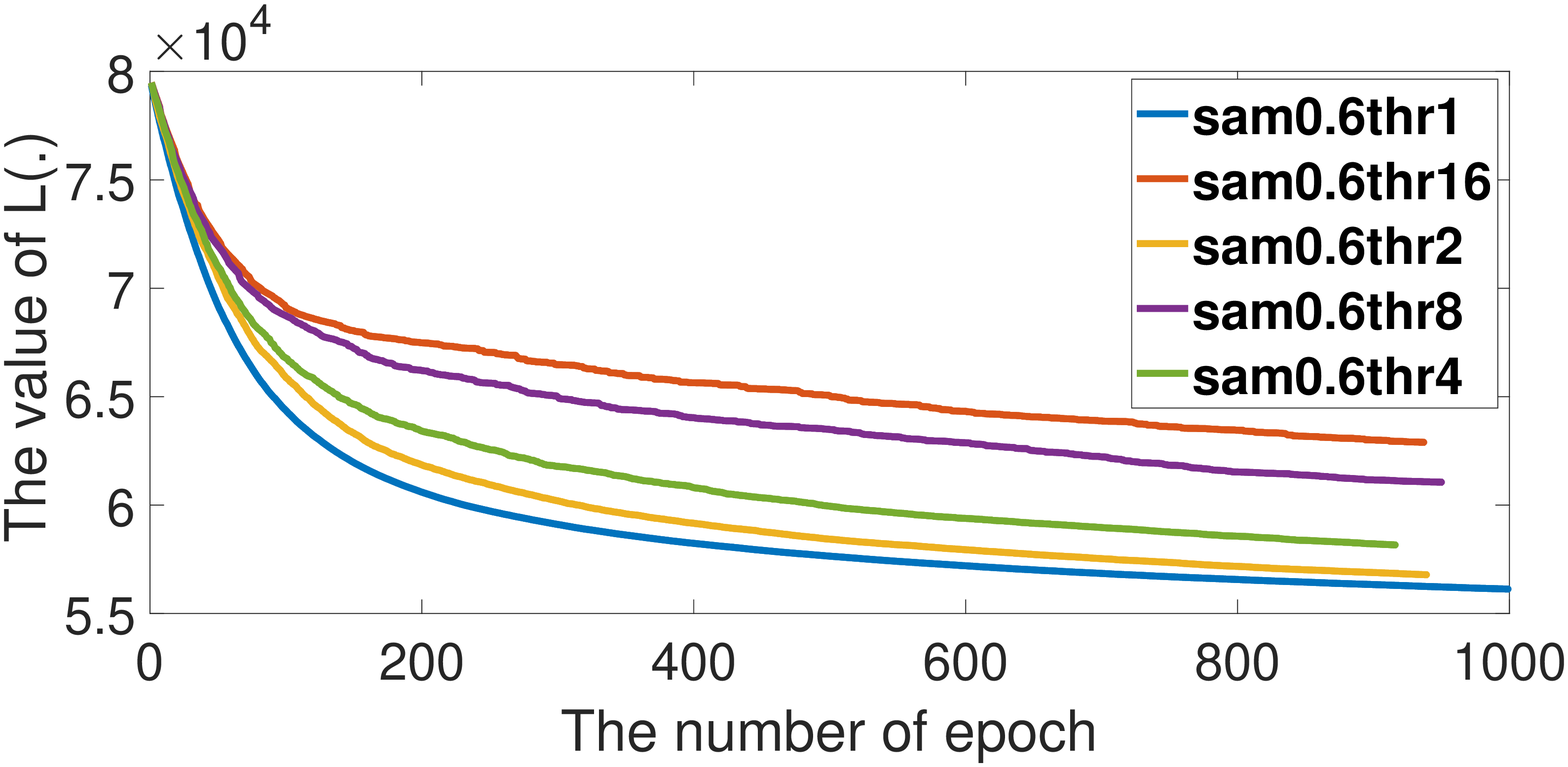}
 	}
 	\subfigure[sampling rate of 0.8 ]{
 		\includegraphics[width=0.47\textwidth]{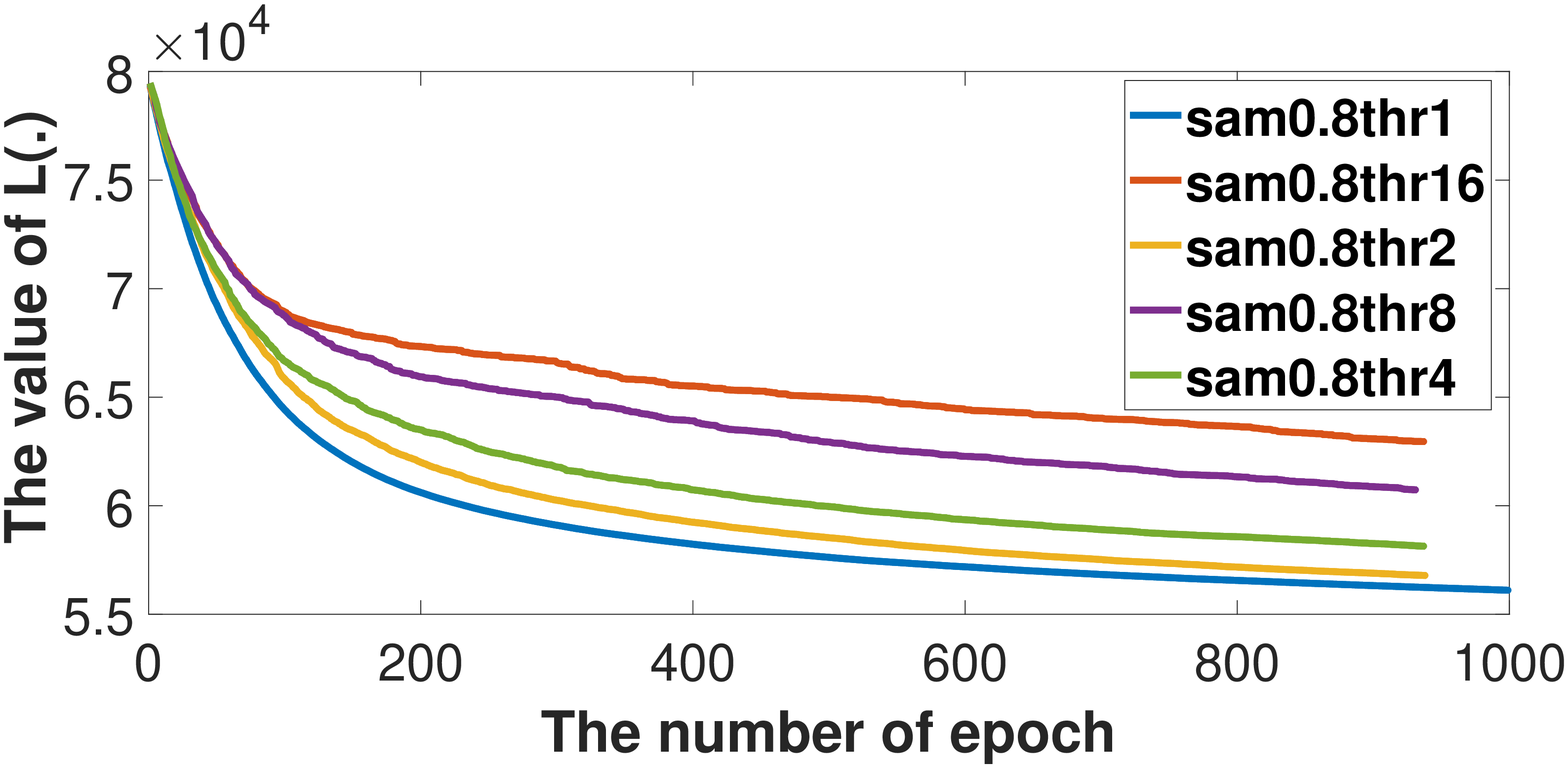}
 		
 	}
 	\caption{Asynch-SGBDT with different number of workers and the same sampling rate using the Higgs dataset}	
 	\label{different_thread_HIGGS}
 \end{figure}

\begin{figure}[tb]
 	\centering
 	\subfigure[1 worker ]{
 		
 		\includegraphics[width=0.47\textwidth]{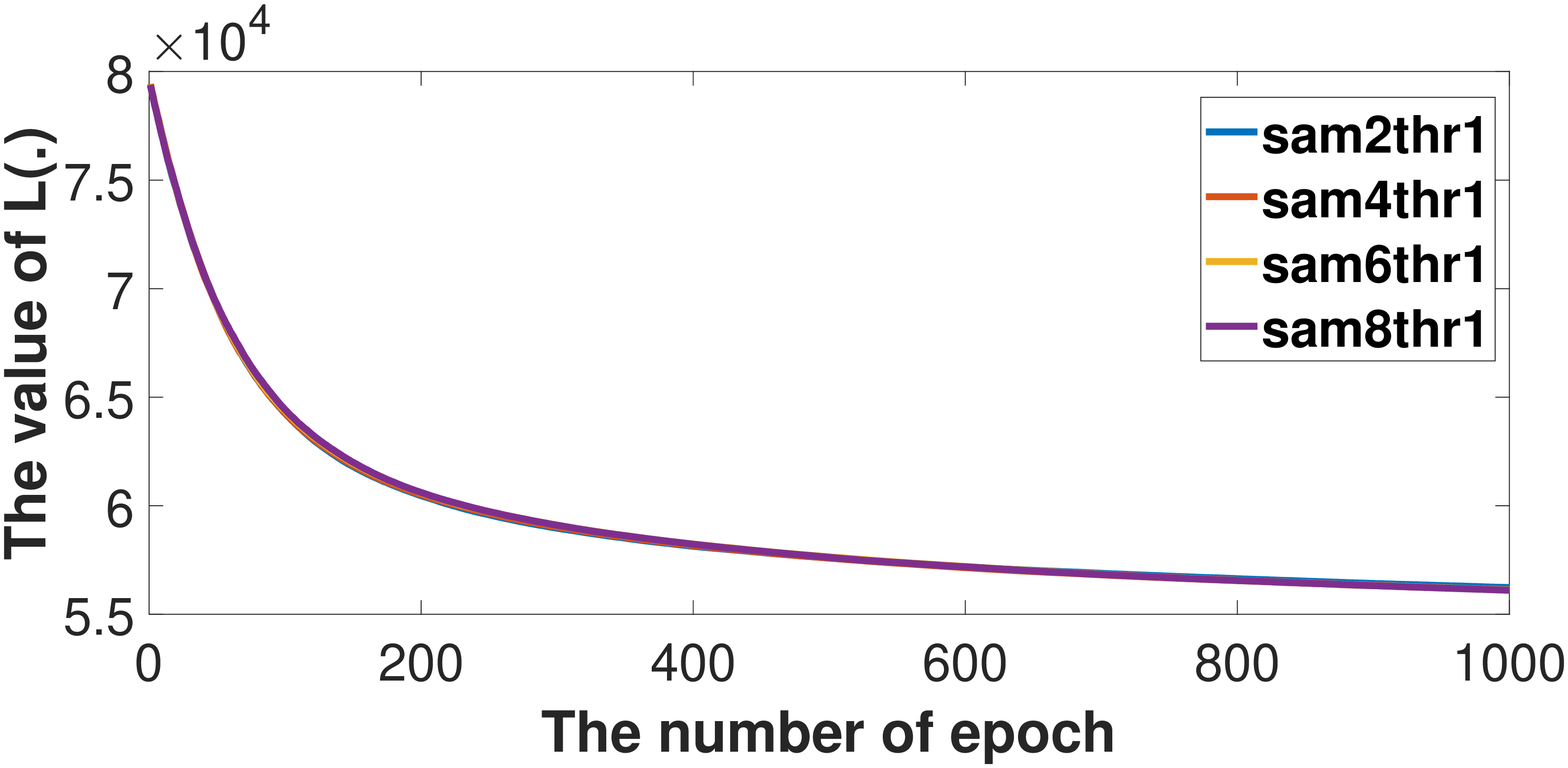}
 		
 	} 
 	\subfigure[4 workers]{

 		\includegraphics[width=0.47\textwidth]{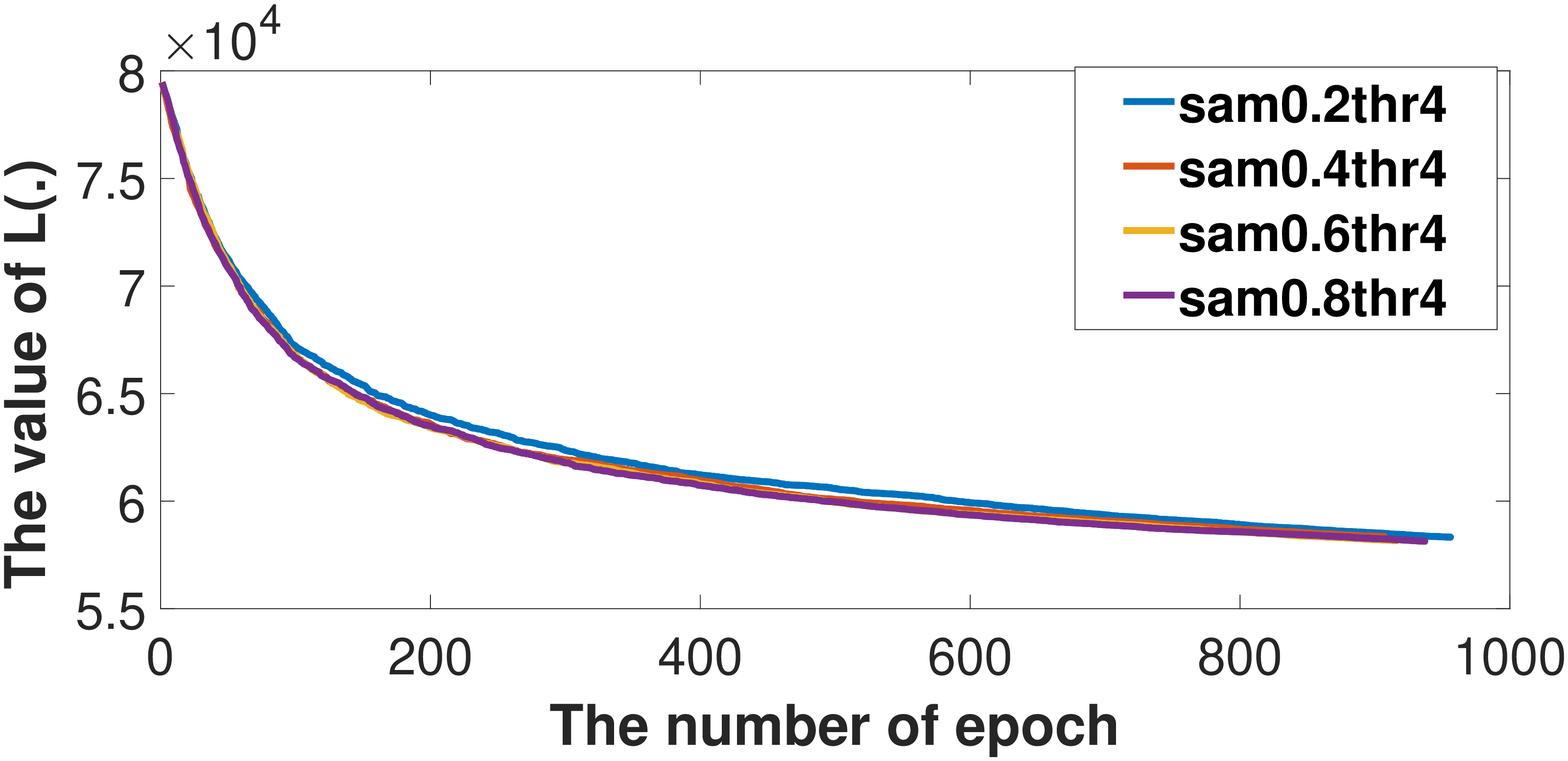}

 	}
 	\subfigure[16 workers ]{
 		
 		\includegraphics[width=0.47\textwidth]{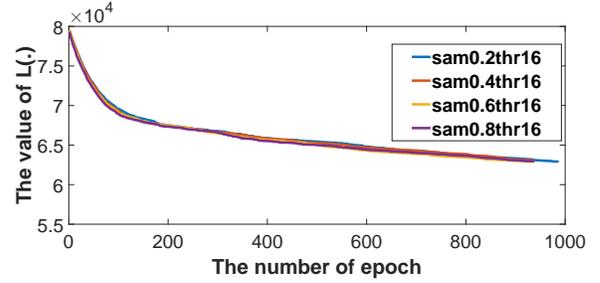}

 	}
 	\caption{Asynch-SGBDT with different sampling rates and the same number of workers using the Higgs dataset}	
 	\label{different_sam_HIGGS}
 \end{figure}

 \textbf{Convergence Speed and Output} The Higgs experimental results are shown in Figures \ref{different_thread_HIGGS} and \ref{different_sam_HIGGS}.  Figure \ref{different_thread_HIGGS} shows that under the fixed sampling rate settings, the more workers we use, the slower the convergence speed.
 
 \begin{figure}[!tb]
 	\centering
 	\subfigure[sampling rate of 0.2 ]{
 		
 		\includegraphics[width=0.47\textwidth]{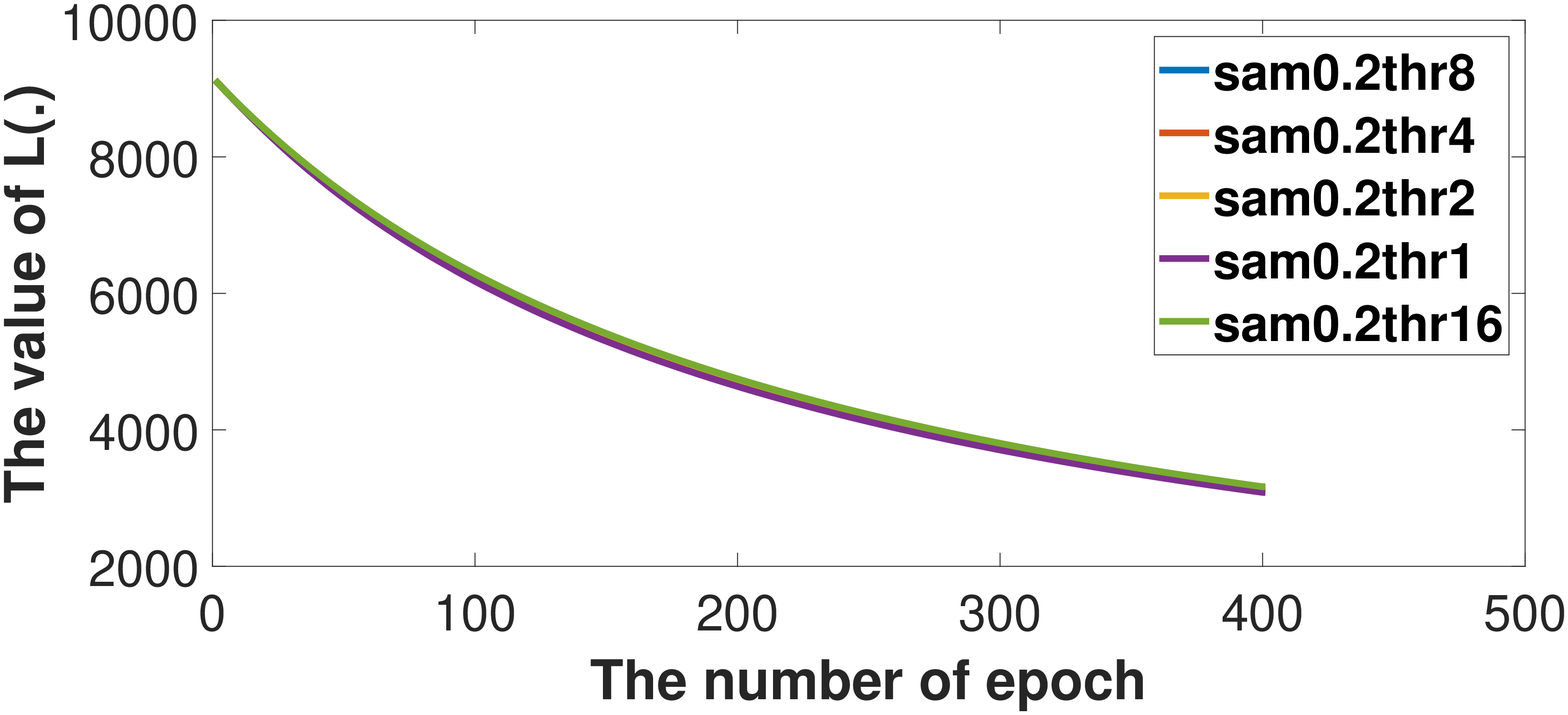}
 		
 	} 
 	\subfigure[sampling rate 0.4 ]{

 		\includegraphics[width=0.47\textwidth]{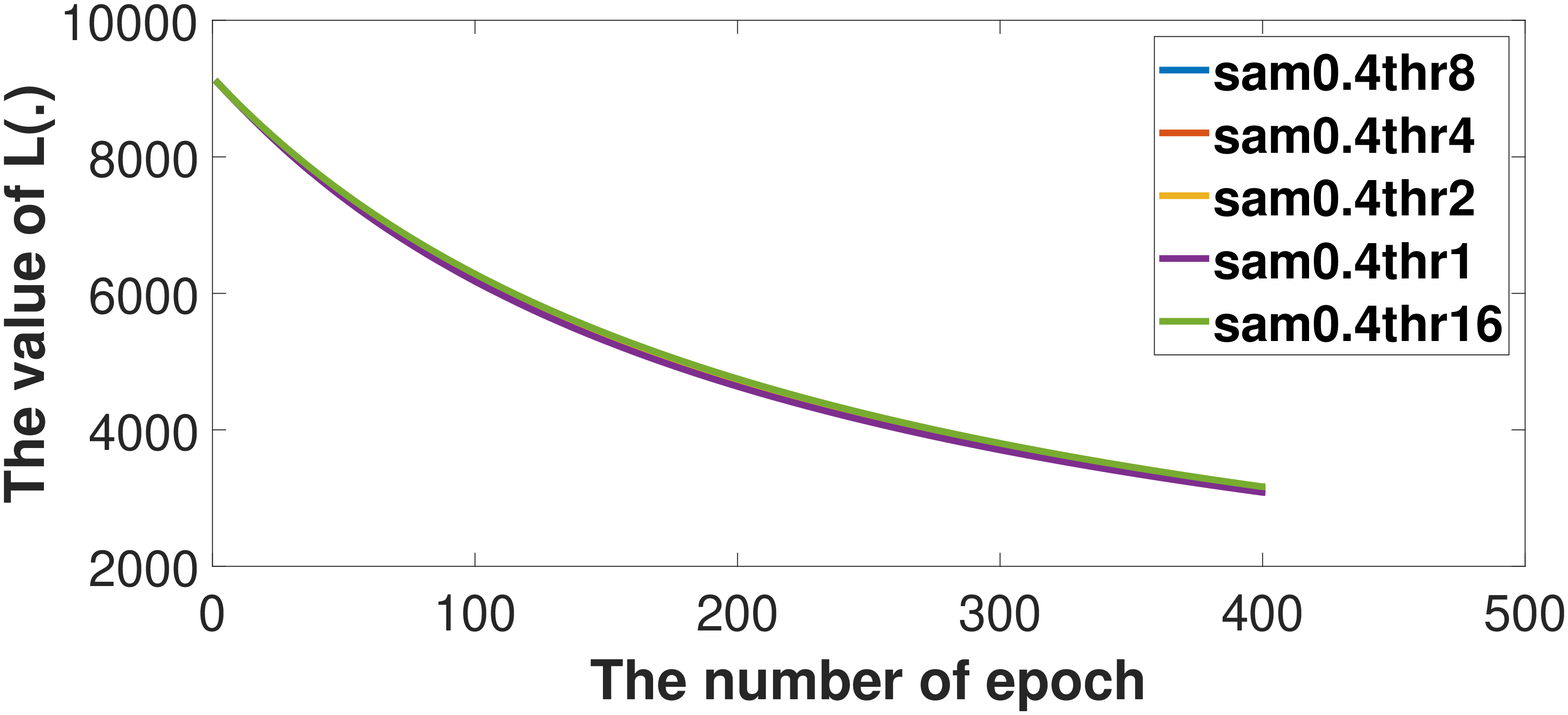}

 	}
 	\subfigure[sampling rate 0.6 ]{
 		
 		\includegraphics[width=0.47\textwidth]{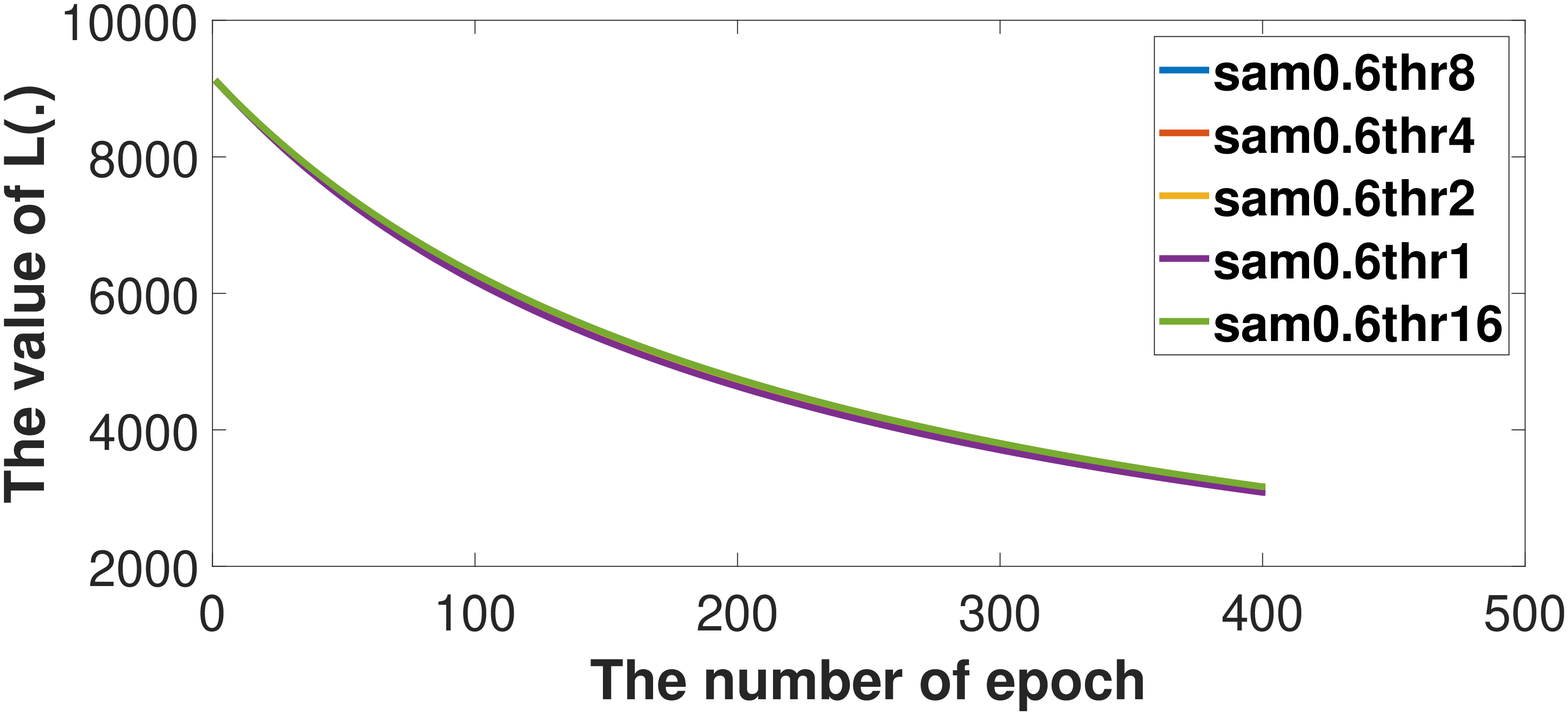}

 	}
 	\subfigure[sampling rate of 0.8 ]{

 		\includegraphics[width=0.47\textwidth]{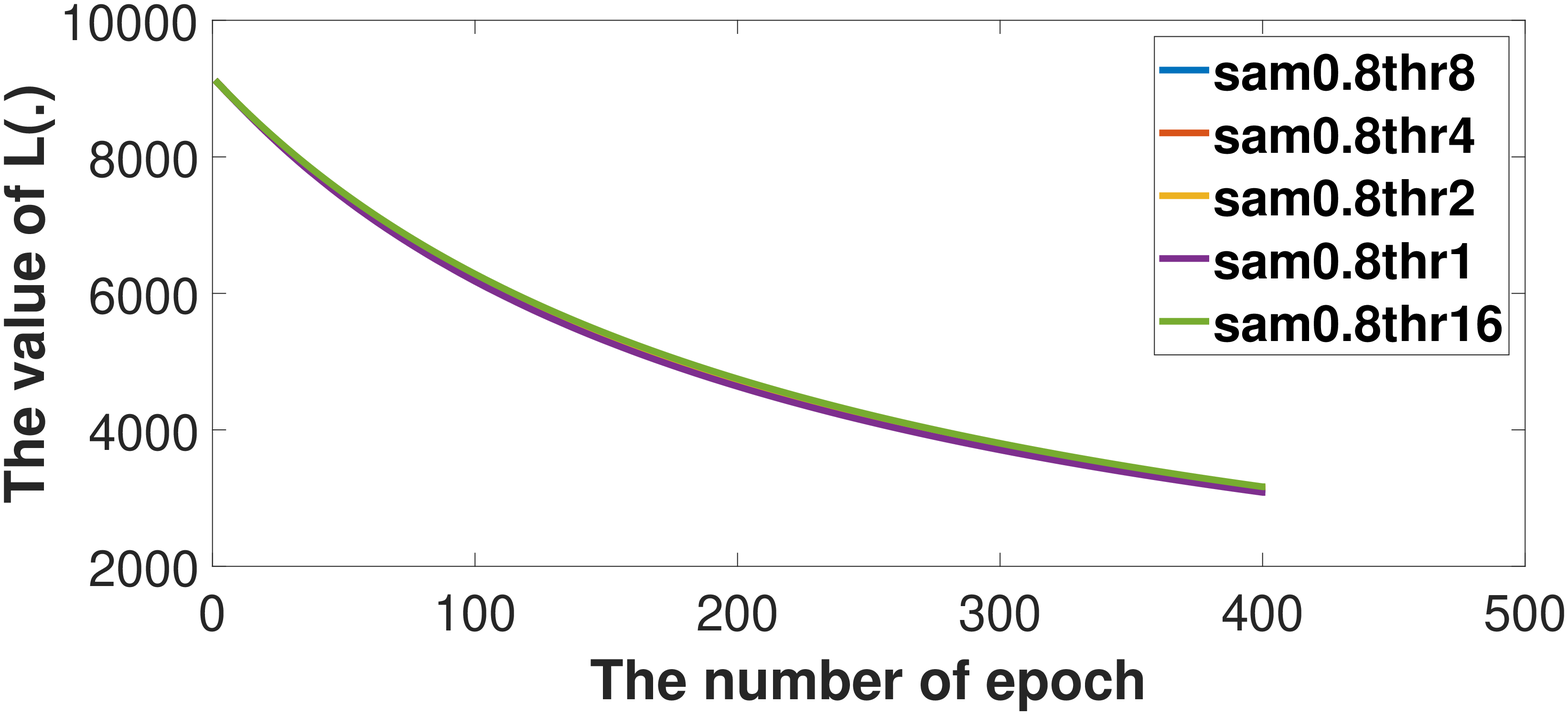}
 		
 	}
 	\caption{Asynch-SGBDT with a different number of workers and the same sampling rate using the real-sim dataset}	
 	\label{different_thread_real}
 \end{figure}

\begin{figure}[tb]
 	\centering
 	\subfigure[1 worker ]{
 		
 		\includegraphics[width=0.47\textwidth]{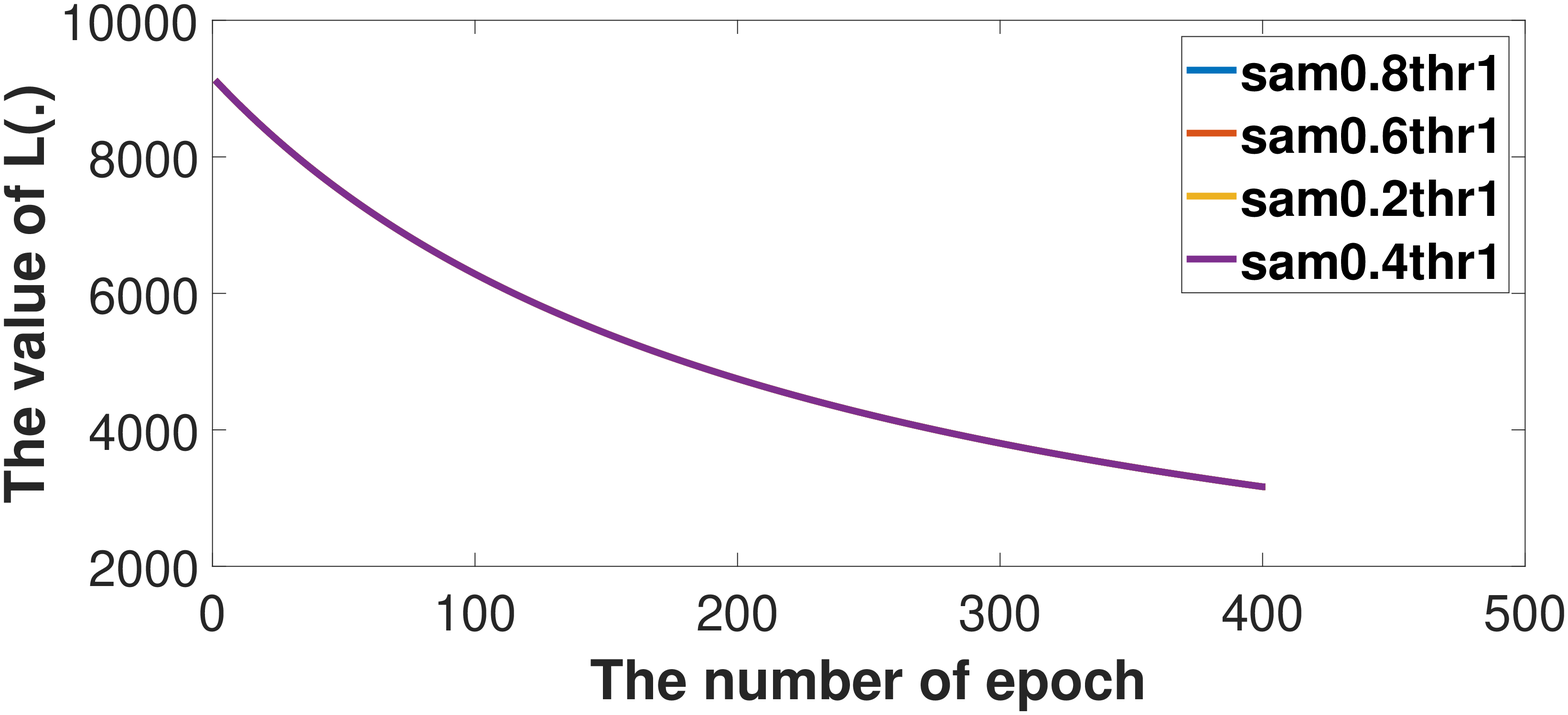}
 		
 	} 
 	\subfigure[4 workers ]{

 		\includegraphics[width=0.47\textwidth]{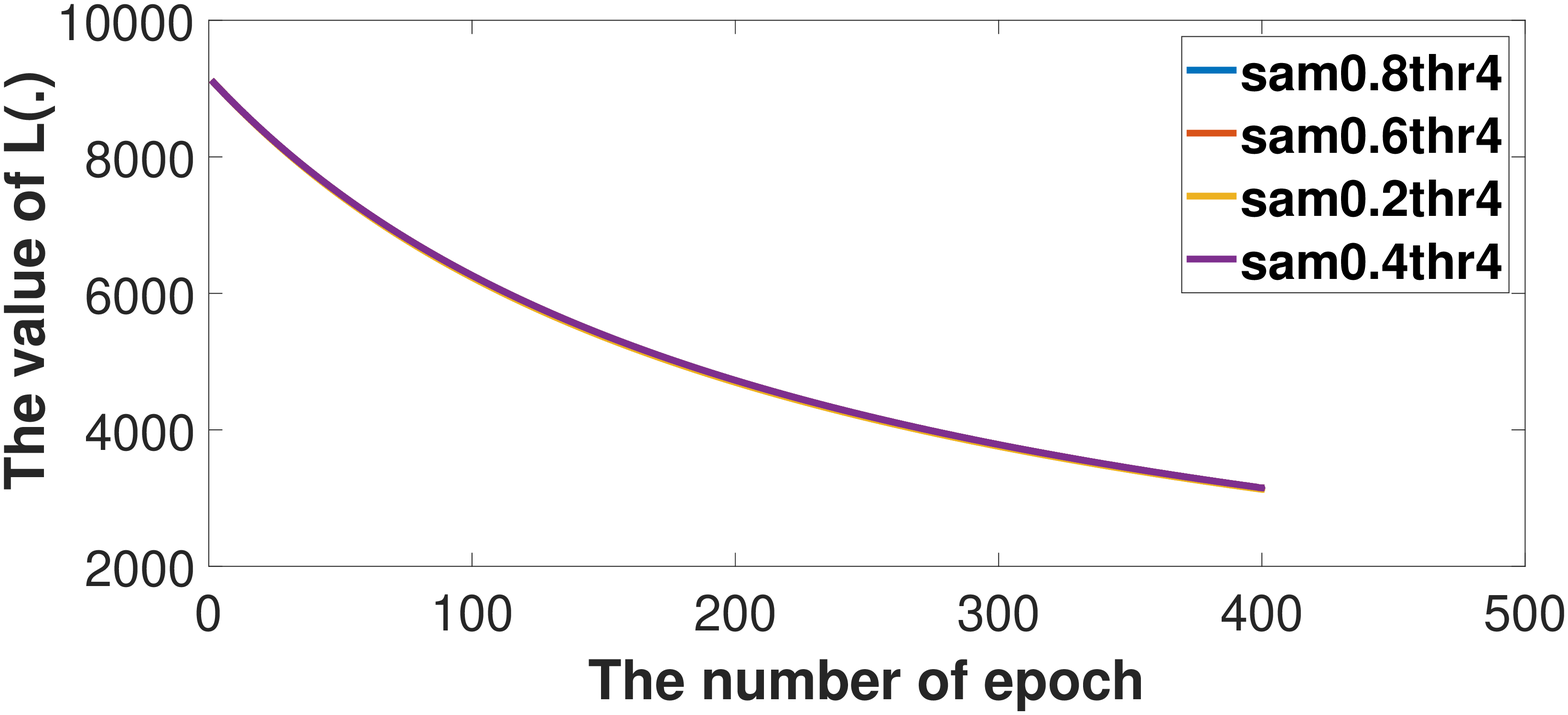}

 	}
 	\subfigure[16 workers ]{
 		
 		\includegraphics[width=0.47\textwidth]{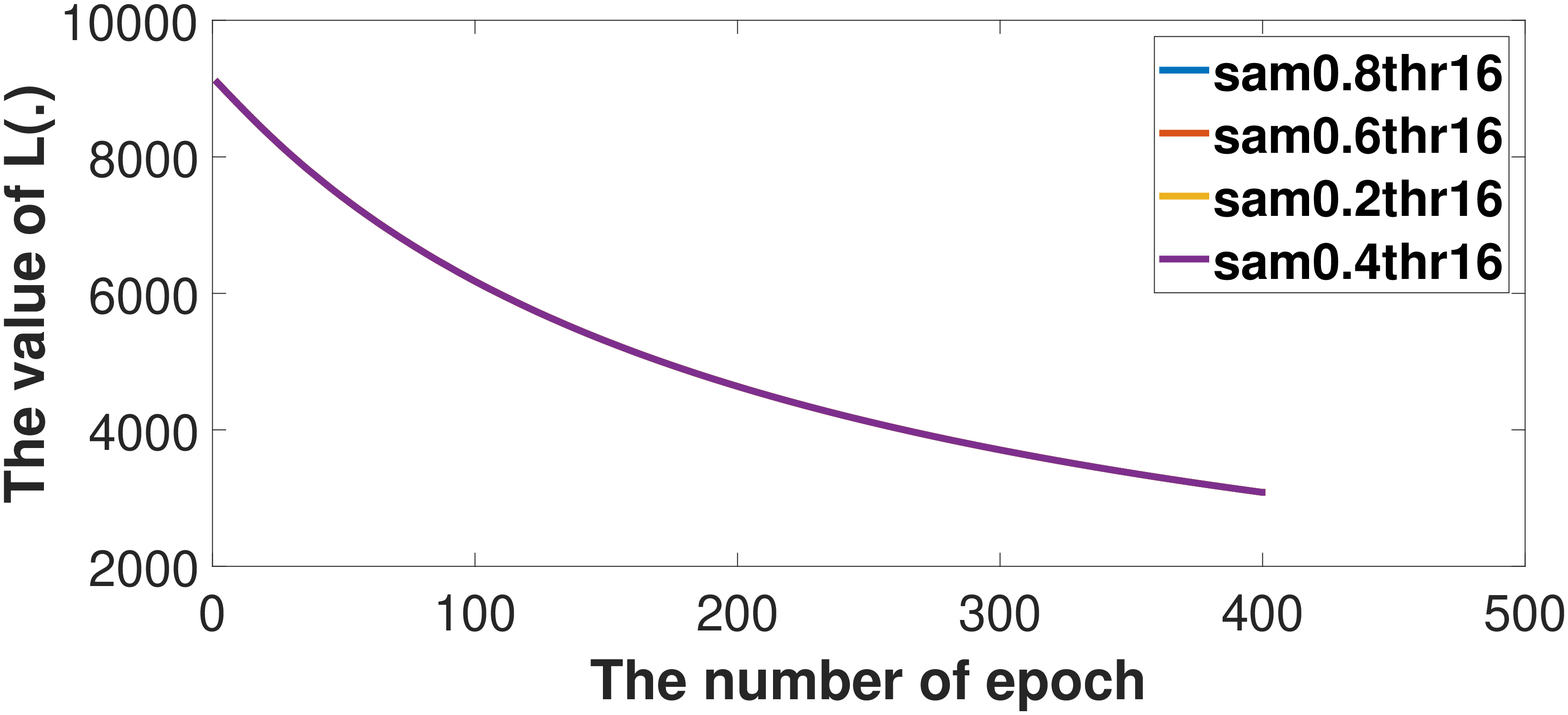}

 	}
 	\caption{Asynch-SGBDT with different sampling rates and the same number of workers using the real-sim dataset}	
 	\label{different_sam_real}
 \end{figure}

 The real-sim experimental results are shown in Figures \ref{different_thread_real} and \ref{different_sam_real}. Figure \ref{different_thread_real} shows that under the fixed sampling settings, the more workers we use, the slightly slower the convergence speed. Experimental results shows that the speedup rises linearly with the increase in the number of workers in unlimited network resource condition. These experimental results are described by conclusion 2 in section 5.2, where the more workers we use, the slower the convergence.
 
 Because of our experimental fixed step length settings, our experimental results are difficult to match conclusion 4 in section 5.2 directly. However, Figures \ref{different_thread_real} and  \ref{different_sam_real} show that the convergence speeds are almost the same when the sampling rates are within a specified range and the step length settings are the same under different experimental settings. A larger step length setting corresponds to a faster convergence speed. Thus, if we set a larger step length for the small sampling experiments, the convergence speed of the small sampling rate would be faster than that of the current experimental setting. The above experimental results and analysis match conclusion 4 in section 5.2 indirectly.
 
 Above figures show that sampling rates between 0.2 and 0.8 exert a slight effect on the convergence speed in this dataset.

 It is worth mentioning that  Figures \ref{different_thread_HIGGS} and  \ref{different_sam_HIGGS} are benchmark experiments for the dataset is lacking sample diversity. We discuss this problem in the next section.



 \textbf{Sensitivity and Scalability} The Higgs and real-sim experiments show that asynch-SGBDT exhibits different sensitivities to asynchronous parallel methods under different settings and datasets, as shown in Figure \ref{different_sam_HIGGS} and Figure \ref{different_sam_real}.
 
 In the Higgs experiments, the dimension of the sample vector and the range of a feature value are relatively small, which leads to diversity in the Higgs dataset being relatively small. From a mathematical perspective, this result means that the observed value vector of  $\mathbf{Q'}$ is almost equal to $[1,1,...,1]$ in every sampling process. $\Delta$ and $\rho$ would be large to this situation. Additionally, the number of leaves on each tree is small, which is also caused by the low dimension of the samples. In this case, similar samples are treated as the same sample, which would reduce the diversity of the samples in the dataset. Low diversity in the dataset increases the sensitivity of the algorithm. Therefore, asynch-SGBDT would be sensitive to the number of workers using the Higgs dataset in our experimental settings.
 
 In the real-sim experiment, the dimension of the samples is large, which increases the diversity of the samples in the dataset and helps to reduce the algorithm sensitivity to delay. Proposition 1 and the experimental results show that asynch-SGBDT would be insensitive to the number of workers using the real-sim dataset under our experimental setting. 
 
 The different sensibilities to the change in the number of workers between the Higgs dataset and real-sim dataset match conclusions 5 and 6 in section 5.2.
 
 \begin{figure}[tb]
 	\centering
 	\subfigure[sampling rate of 0.6]{
 		
 		\includegraphics[width=0.47\textwidth]{6rate_higgs}

 	} 
 	\subfigure[ sampling rate of $5*10^{-6}$]{

 		\includegraphics[width=0.47\textwidth]{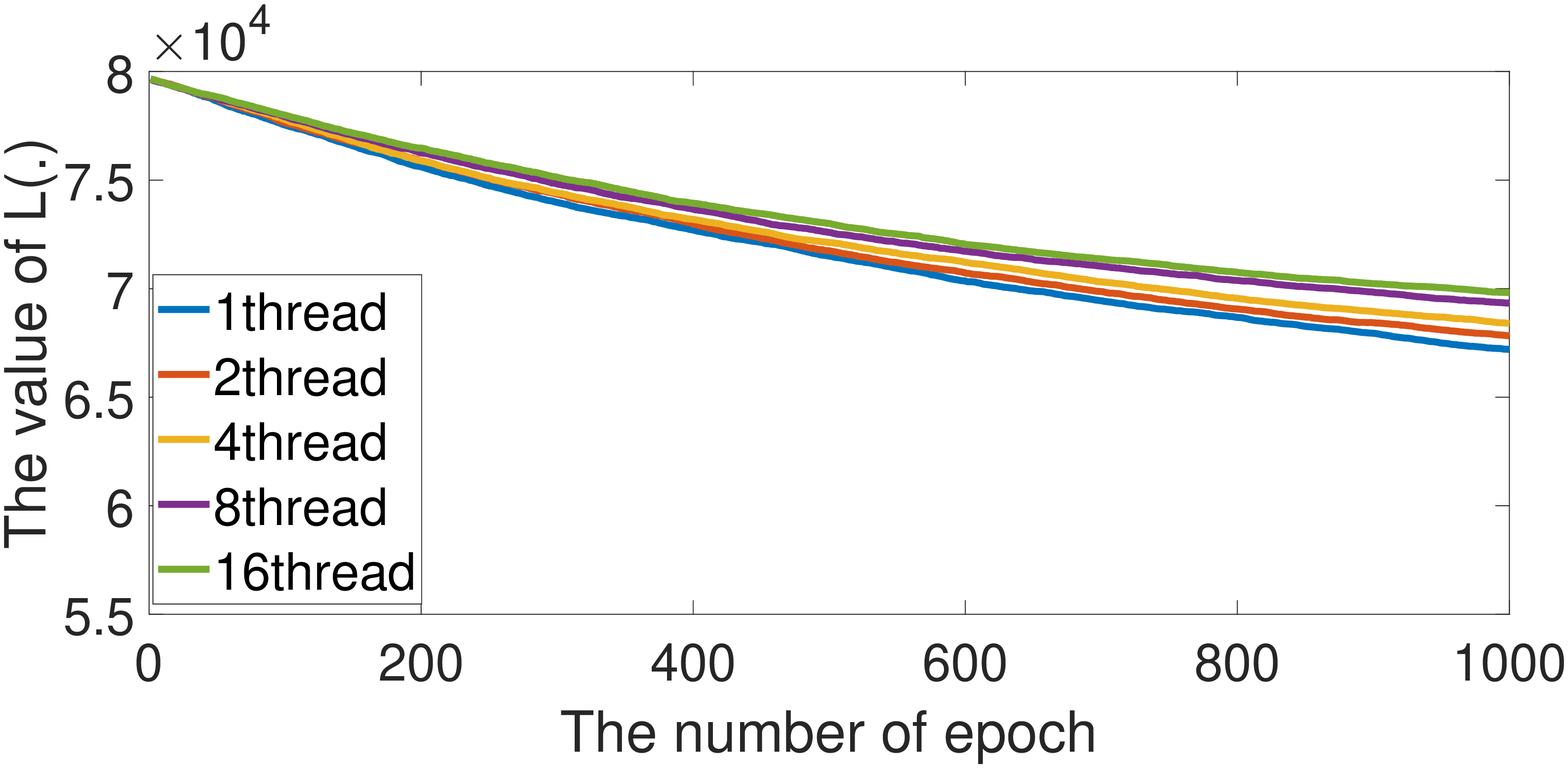}
 		
 	}
 	\caption{Sensitivity between the normal sampling rate and extremely small sampling rate }	
 	\label{6vsverysamll}
 \end{figure}
 \begin{figure}[tb]
 	\centering
 	\subfigure[real-sim]{
 		
 		\includegraphics[width=0.47\textwidth]{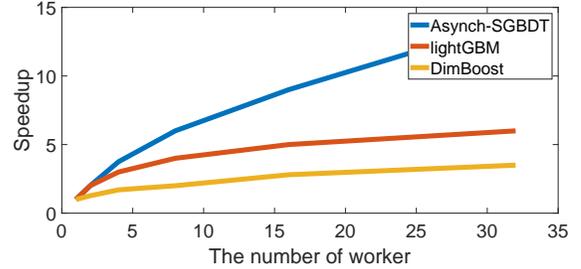}

 	} 
 	\subfigure[E2006-log1p]{

 		\includegraphics[width=0.47\textwidth]{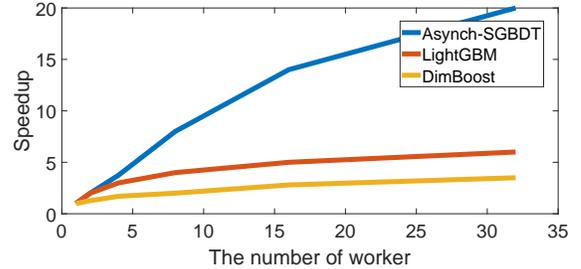}
 		
 	}
 	\caption{Speedup for asynch-SGBDT and LightGBM }	
 	\label{speedup}
 \end{figure}
 
 In addition to the above experiments, we also conducted an experiment using an extremely small sampling rate (sampling rate = 0.000005, which means we  use approximately 500 samples on average in each sampling subdataset). The baseline experiment uses a normal sampling rate (sampling rate = 0.6). The result is shown in Figure \ref{6vsverysamll}. The small sampling rate, which produces a small subdataset, reduces the sample diversity in the sampling subdataset. 
 Small sample diversity in the sampled dataset would help reduce $\Delta$ and $\rho$. This experiment shows that a small sampling rate would help reduce the sensitivity with the help of reducing $\Delta$ and $\rho$. However, an extremely small sampling rate would decrease the convergence speed because the subdataset is too small, which would cause the GBDT trees to be distorted.   The experimental results in Figure \ref{6vsverysamll} match conclusions 1 and 3 in section 5.2.
 \subsection{Efficiency experiments}
 
 \subsubsection{Experiment Settings} This part will introduce our efficiency experiments setting.
 
 \textbf{Algorithm Setting:} In the real-sim and  E2006-log1p experiments, we built 400 trees in total, and each tree had a maximum of 400 leaves.   We randomly sampled 80\% of features in the experiments to build a tree at each building tree sub-step. The step length ($v$) in the experiments was fixed at 0.01. To gain clear experimental results, we set all sampling rates ($R_{i,j}$) to be the same as $0.8$. We use gradient step in LightGBM boosting training process.
 
 \textbf{Code Setting:} We adapted  LightGBM \cite{NIPS2017_6907} on ps-lite\cite{li2014scaling} as asynch-SGBDT code. The performance of feature parallel method in LightGBM is the benchmark in our experiment. Above codes are shared the same tree building step codes ($treelearner$ code in LightGBM). To make a comparison on the same baseline, we only use one OpenMP thread in LightGBM's building tree step, because ps-lite have to use extra thread maintain parameter server network.
 
 \textbf{Computing Environment:} All of the experiments are conducted on Era supercomputing environment. We use TCP/IP network in Era supercomputing as connection methods. TCP/IP network is consist of Intel(R) I350 Gigabit Network Connection.
 
 \textbf{Benchmark setting:} Our benchmark is LightGBM, which is reported as state of the art high performance parallel GBDT implementation\cite{NIPS2017_6907}, and some blogs, kaggle and GitHub issue gain the conclusion that LightGBM is almost seven times faster than XGBOOST and is a much better approach when dealing with large datasets. We also gain the Dimboostdata from paper \cite{inproceedings}, which is a upgrade version of tencentboost\cite{Jiang2017TencentBoost}. Both of the above synchronous parallel GBDT frameworks are state of the art synchronous parallel works. These benchmarks will strongly show that our algorithm is better than synchronous parallel GBDT algorithms.

 \textbf{Measure method:} To make our presentation clearly, we use speedup value as our measure value.

 Because of our experiments code setting, the time costs of 1 worker for asynch-SGBDT and LightGBM are the same. Above fact shows that the speedup comparesion can map to the real time   comparesion in the experiments of LightGBM and asynch-SGBDT.
  
 The data of Dimboost cannot be mapped into real-time for those data are gain from different platform. We offer this data for for referrence use. Actually, Dimboost shows its great advantages on the ability of fast building decision tree in real time.
 
 \subsubsection{Analysis of Experimental Results}
 
 Figure \ref{speedup} shows the speedup between asynch-SGBDT, Dimboost and LightGBM. Our experiment shows that in real-sim dataset experiment, asynch-SGBDT achieves 14x speedup; in E2006-log1p dataset experiment, asynch-SGBDT achieves 20x speedup using 32 workers. However, feature parallel method used in LightGBM is not effective as asynch-SGBDT: in real-sim dataset and E2006-log1p dataset, LightGBM achieves 6x-7x speedup using 32 machines.  In Dimboost experiments, in real-sim dataset and E2006-log1p dataset, Dimboost achieves 4x-6x speedup using 32 machines.

 Asynch-SGBDT and LightGBM show a great difference in speedup. Especially with the increase of the number of machines or workers, the gap is expanded. This phenomenon is caused by 1. The node performance inconsistency in a cluster. In the synchronous parallel algorithm, the algorithm does not continue the process until all nodes reach the barrier. The increase of the number of machines would exacerbate the cost of synchronous operation. 2. Asynch-SGBDT is robust against network instability and the computing time and communication time is overlapped in asynch-SGBDT. 3.  In Asynch-SGBDT, whole building tree process is running parallel. In LightGBM, only collection feature information process can be implemented in a parallel method. Collection feature information process is just a part of build tree process.

 In Dimboost, tencentboost, the allgather operation is done by parameters server. Parameter server's allgather is a centralization operation. The burden of the server is the key for scalability. Thus, with the number of worker's increasing, the high cost of communication is unavoidable, which leads to low scalability. 
 
 Asynch-SGBDT experiments using different datasets also show a great difference in speedup. This phenomenon is caused by the time of building tree step in real-sim dataset is small. Eq.\ref{max_speedup} shows that in the case of real-sim, 16 to 32 worker is close to the max number of the worker in asynch-SGBDT.

 %
 %
 %
 %

 \section{Conclusion }
 We propose a novel algorithm, asynch-SGBDT, that provides good compatibility for the parameter server. The theoretical analyses and experimental results show that a small iteration step, small sampling rate, large number of workers, and high-dimensional sparsity of the sample datasets lead to a fast convergence speed and high scalability. Specifically, asynch-SGBDT reaches a high speedup when asynch-SGBDT uses high-dimensional sparse datasets.  
 

\bibliographystyle{IEEEtran}
\bibliography{IEEEabrv,mybibfile_gbdt}

\section*{Appendix}
\subsection{Iteration Step}
To make our analysis more general, we will use following iteration step to describe our algorithm

\begin{align}
\mathbf{F}^{j+1}&=\mathbf{F}^{j}-vV(V^TV)^{-1}V^TL'^{k(j)}_{random}\notag \\
&=\mathbf{F}^{j}-vV(V^TV)^{-1}V^TPL_{base}'^{k(j)}
\end{align}

where $L_{base}=(\ell_1,\ell_2,...,\ell_N)$.
The matrix $V$ is defined as the projection matrix in Sun's work \cite{Sun2014A} and $P$ is the diagonal  random variable matrix: $P = diag(\sum_{j=1}^{m_1}\frac{Q_{1,j}}{R_{1,j}},\sum_{j=1}^{m_2}\frac{Q_{2,j}}{R_{2,j}},...,\sum_{j=1}^{m_N}\frac{Q_{N,j}}{R_{N,j}})$ and $P_{fix} = diag(m_1,m_2,...,m_N)$. Thus, $\mathbb{E}P = P_{fix}$

We can summarize the following relationship:
\begin{equation*}
P_{fix}L'_{base} = \mathbf{G} = \mathbb{E}PL'_{base}=\mathbb{E}L'_{random}
\end{equation*}

In following presentation, we let $A=V(V^TV)^{-1}V^T$.

\subsection{New and redefined note}
To make readers have a clear comparison with special case, we define some new note and redefine some old notes in the body of paper. In appendix, all notes are based on this subsection. 

$\lambda$ is the Lipschitz constant for $\ell(.)$. $L(.)$ is a strong convex function with modulus $c$. For $\mathbf{F} \in \mathbb{R}^N$ and every observed value vector corresponding to $\mathbf{Q}$, $\|AL'_{random}\|<M$. 
 
$\Omega$  represents the maximum number of different non-zero component for the vector $ AL'_{random} $.   
 
$\rho$ represents the probability that two vector multiplication of $ AL'_{random} $ is zero.
 
$\zeta$ represents the maximum number of non-zero component for the vector $ AL'_{random} - L'_{random} $. This value can measure the tortuosity of decision tree to $L'_{random}$. Apparently, when the tree has more leaves,  $\zeta$ is close to zero.

\subsection{Assumption}
It is normal to use following assumptions.
\subsubsection{$c$-strong convex}

The $f(\cdot)$ is $c$-strong convex function, i.e.
\begin{equation}
f(F^{j+1})-f(F^{j})+\frac{c}{2}\left\| F^{j+1} - F^{j}  \right\|^2 \le (F^{j+1} - F^{j} )^Tf'(F^{j+1})
\end{equation}

Let $F^*$ be the minimal of $\ell$, and we have
\begin{equation}
\frac{c}{2}\left\| F^{j} - F^{*}  \right\|^2 \le (F^{j} - F^{*} )^Tf'(F^{j})
\end{equation} 

In our paper, $L(\mathbf{F})$ and $\ell(\cdot,\cdot)$ are $c$-strong convex function.

\subsubsection{$\lambda$-Lipschit function}
The $f(\cdot)$ is $\lambda$-Lipschit  function, i.e.

\begin{equation}
 \left\|  f'(F^{j+1}) -  f'(F^{j}) \right\| \le \lambda \left \| F^{j+1} - F^{j} \right\|
\end{equation}


In our paper,  $\ell(\cdot,\cdot)$ are  $\lambda$-Lipschit  functions.%
 $L(\mathbf{F})$  are  $m_{max}N$-strong Lipschit smoothness functions.

\subsubsection{Bound of gradient's norm}

In this paper, we need to set the bound of gradient's norm.
\begin{equation*}
\left\| \ell' \right\| \le \phi
\end{equation*}
Thus, we define the bound of gradient of $L(\cdot)$, $M$, and we gain the following equations.

\begin{equation*}
 \Omega m^2_{max} \phi^2 \ge M^2 \ge L_{base}'^TP^TAPL_{base}'  
\end{equation*}



\subsubsection{The sample's similarities in one GBDT's leaf}

For the decision tree is the classifier which classify sample via their space position. For the samples $\{(x_i,y_i)\}^{\sum{j=1}{N}m_j}$ in one leaf, we have following equation:

\begin{equation}
\mathop{max}_{i,j \in leaf_k}\left\| x_i - x_j \right\| \le \delta
\end{equation}

\subsection{Proof}

\subsubsection{section 1: basic transformation}
The Proof begin at iteration step
\begin{equation*}
\mathbf{F}^{j+1} = \mathbf{F}^{j} - vA_{k(j)}P_{k(j)}L_{base}'(\mathbf{F}^{k(j)})
\end{equation*}

The minimum of $L(\mathbf{F})$ is $\mathbf{F^*}$. And Let $\beta = APL_{base}'(\mathbf{F})$.

\begin{align*}
&(\mathbf{F}^{j+1} -\mathbf{F}^{*})  = ( \mathbf{F}^{j} -\mathbf{F}^{*}) - vAPL_{base}'(\mathbf{F}^{k(j)})\\
&(\mathbf{F}^{j+1} -\mathbf{F}^{*})^2  = ( \mathbf{F}^{j} -\mathbf{F}^{*})^2 -2v( \mathbf{F}^{j} -\mathbf{F}^{*})^T \beta_{k(j)}
\\
&+ v^2\beta_{k(j)}^2\\
&(\mathbf{F}^{j+1} -\mathbf{F}^{*})^2  = ( \mathbf{F}^{j} -\mathbf{F}^{*})^2 + v^2
\beta_{k(j)}^2\\
&-2v( (\mathbf{F}^{j} - \mathbf{F}^{k(j)})^T\beta_{j} + (\mathbf{F}^j - \mathbf{F}^{k(j)})^T(\beta_{j}-\beta_{k(j)}) + (\mathbf{F}^{k(j)}\\
& - \mathbf{F}^* )^T\beta_{k(j)})
\end{align*}

we will find the lower bound of 
\begin{equation}
(\mathbf{F}^{j} - \mathbf{F}^{k(j)})^T\beta_{j} + (\mathbf{F}^j - \mathbf{F}^{k(j)})^T(\beta_{j}-\beta_{k(j)}) + (\mathbf{F}^{k(j)} - \mathbf{F}^* )^T\beta_{k(j)}
\end{equation}
This part is consist of first item $(\mathbf{F}^{j} - \mathbf{F}^{k(j)})^T\beta_{j}$, 2nd item $ (\mathbf{F}^j - \mathbf{F}^{k(j)})^T(\beta_{j}-\beta_{k(j)})$ and 3rd item $ (\mathbf{F}^{k(j)} - \mathbf{F}^* )^T\beta_{k(j)}$.

\subsubsection{section 2: Extra Lemma}
Before continue proof, we need following extra lemma to support our next proof.

\textbf{Lemma 1} \begin{align*}
	&\mathbb{E}(\mathbf{F}^j-\mathbf{F}^i)APL_{base}'(\mathbf{F^j})\\
	\ge &\mathbb{E}( L(\mathbf{F}^j) - L(\mathbf{F}^i) + \frac{c}{2}\left\| \mathbf{F}^j-\mathbf{F}^i    \right\|^2) - C_1 \left\| \mathbf{F}^j-\mathbf{F}^i    \right\|
	\end{align*}
	
	\begin{proof}
		\begin{align*}
			&\mathbb{E}	(\mathbf{F}^j-\mathbf{F}^i)APL_{base}'(\mathbf{F^j})\\
			& = \mathbb{E} (\mathbf{F}^j-\mathbf{F}^i)P_{fix}L_{base}'(\mathbf{F^j})+	\mathbb{E}(\mathbf{F}^j-\mathbf{F}^i)(AP-P_{fix})L_{base}'(\mathbf{F^j})			
		\end{align*}
		For matrix multiplication is Linear operation. Thus,
		\begin{align*}
		&\mathbb{E}	(\mathbf{F}^j-\mathbf{F}^i)APL_{base}'(\mathbf{F^j})\\
		=&\mathbb{E}(\mathbf{F}^j-\mathbf{F}^i)P_{fix}L_{base}'(\mathbf{F^j})+	\mathbb{E}(\mathbf{F}^j-\mathbf{F}^i)(\mathbb{E}AP-P_{fix})L_{base}'(\mathbf{F^j})\\
		\ge &\mathbb{E}(  L(\mathbf{F}^i) - L(\mathbf{F}^j) + \frac{c}{2}\left\| \mathbf{F}^j-\mathbf{F}^i    \right\|^2 \\
		& - \left\|\mathbf{F}^j-\mathbf{F}^i \right\|\left\|  (AP-P_{fix})L_{base}'(\mathbf{F^j})  \right\|)
		\end{align*}

		for any observation vector $\mathbf{Q}$, the vector $ (AP-P_{fix})L_{base}'(\mathbf{F^j}) $  always is the vector like
		\begin{align*}
		&(AP-P_{fix})L_{base}'(\mathbf{F^j})\\
		&= \mathbb(...,m_s \frac{\sum_{ (x_r,y_r) \in leaf_J }m_r\ell'(y_r,F_r)}{\sum_{ (x_r,y_r) \in leaf_s }m_r}  -m_s\ell'(x_s,F_s) ,...)^T
		\end{align*} 
		
		Thus, we can gain it norm upper bound that:
		
		\begin{align*}
		&\left\| m_s\frac{\sum_{ (x_r,y_r) \in leaf_J }m_r\ell'(y_r,F_r)}{\sum_{ (x_r,y_r) \in leaf_s }m_r}  -m_s\ell'(x_s,F_s)\right\|\\
		=&m_s\left\|    \frac{\sum_{ (x_r,y_r) \in leaf_J }m_r(\ell'(y_r,F_r)-\ell(y_s,F_s))}{\sum_{ (x_r,y_r) \in leaf_s }m_r}    \right\|\\
		\le&\frac{m_s}{\sum_{ (x_r,y_r) \in leaf_s }m_r} \left\|  \sum_{ (x_r,y_r) \in leaf_J }(m_r(\ell'(y_r,F_r)-\ell'(y_s,F_s)))  \right\|\\
		\le&\frac{m_s}{\sum_{ (x_r,y_r) \in leaf_s }m_r} \sum_{ (x_r,y_r) \in leaf_J } \left\|   \lambda m_r\left\|F_r - F_s\right\|        \right\| \\
		\le&m_s\lambda\delta
		\end{align*} 
	
		Thus, $\left\|  (AP-P_{fix})L_{base}'(\mathbf{F}^j)  \right\| \le \lambda\delta m_{max}\sqrt{\zeta}$ 
		
		And $	\mathbb{E}(\mathbf{F}^j-\mathbf{F}^i)APL_{base}'(\mathbf{F^j})\ge 	\mathbb{E}( L(\mathbf{F}^j) - L(\mathbf{F}^i) + \frac{c}{2}\left\| \mathbf{F}^j-\mathbf{F}^i    \right\|^2 -\lambda\delta m_{max}\sqrt{\zeta} \left\| \mathbf{F}^j-\mathbf{F}^i    \right\|)$
	\end{proof}

\subsubsection{section 3-first item} 
we will show the lower bound of $(\mathbf{F}^{j} - \mathbf{F}^{k(j)})^T\beta_{j}$.

\begin{align*}
&\mathbb{E}(\mathbf{F}^{j} - \mathbf{F}^{k(j)})^T\beta_{j}\\
\ge&\mathbb{E}(L(\mathbf{F}^j)-L(\mathbf{F}^{k(j)}) +\frac{c}{2}\left\| \mathbf{F}^j - \mathbf{F}^{k(j)}  \right\|^2-\lambda\delta\sqrt{\zeta} \left\| \mathbf{F}^j-\mathbf{F}^{k(j)}    \right\|)\\
=&\mathbb{E}(\sum_{i=k(j)}^{j-1}(L(\mathbf{F}^{i+1} )- L(\mathbf{(F)^i}) ) +\frac{c}{2}\left\| \mathbf{F}^j - \mathbf{F}^{k(j)}  \right\|^2\\
&-\lambda\delta m_{max}\sqrt{\zeta} \left\| \mathbf{F}^j-\mathbf{F}^{k(j)}    \right\|)
\end{align*}
Here, we have to show the   bound of $\sum_{i=k(j)}^{j-1}(L(\mathbf{F}^{i+1} )- L(\mathbf{(F)^i}) ) $
\begin{align*}
&\mathbb{E}\sum_{i=k(j)}^{j-1}(L(\mathbf{F}^{i+1} )- L(\mathbf{(F)^i}) ) \\
\le&\mathbb{E}\sum_{i=k(j)}^{j-1}(\mathbf{F}^{i+1}-\mathbf{F}^{i})^TA_iP_iL'(\mathbf{F}^j)\\
=&v\mathbb{E}\sum_{i=k(j)}^{j-1}L'(\mathbf{F}^{k(i)})^TP_{k(i)}^TA_{k(i)}^TA_iP_iL'(\mathbf{F}^j)\\
\le&v \tau\rho M^2 
\end{align*}

Thus
\begin{align*}
&\mathbb{E}(\mathbf{F}^{j} - \mathbf{F}^{k(j)})^T\beta_{j}\\
\ge&\mathbb{E}\frac{c}{2}\left\| \mathbf{F}^j - \mathbf{F}^{k(j)}  \right\|^2 - v\tau\rho M^2 -\lambda\delta m_{max}\sqrt{\zeta}\left\| \mathbf{F}^j-\mathbf{F}^{k(j)}    \right\|
\end{align*}

\subsubsection{section 4-second item}
\begin{align*}
&\mathbb{E}(\mathbf{F}^j-\mathbf{F}^{k(j)})^T(\beta_j-\beta_{k(j)})\\
=&v\mathbb{E}\sum_{i=k(j)}^{j-1}A_iP_iL'(\mathbf{F}^{k(j)})(\beta_j-\beta_{k(j)})\\
\ge&-2v\rho\tau M^2
\end{align*}

\subsubsection{section 5- third item}
\begin{align*}
 &(\mathbf{F}^{k(j)} - \mathbf{F}^* )^T\beta_{k(j)}\\
 \ge&\frac{c}{2}\left\|  \mathbf{F}^{k(j)} - \mathbf{F}^*  \right\|^2-\lambda\delta m_{max}\sqrt{\zeta} \left\|  \mathbf{F}^{k(j)} - \mathbf{F}^*  \right\|
\end{align*}

\subsubsection{section 6-combine section 3 to section 5}

\begin{align*}
&\mathbb{E}(\mathbf{F}^{j+1}-\mathbf{F}^*)^2\\
\le&\mathbb{E}((\mathbf{F}^j - \mathbf{F}^*)^2 +v^2M^2-2v(\frac{c}{2}\left\|  \mathbf{F}^{k(j)} - \mathbf{F}^*\right\|^2\\
& -\delta\lambda m_{max}\sqrt{\zeta}\left\|  \mathbf{F}^{k(j)} - \mathbf{F}^* \right\|-2v\rho\tau M^2\\
&+\frac{c}{2}\left\| \mathbf{F}^j - \mathbf{F}^{k(j)}  \right\|^2 - v\tau\lambda\rho M^2 -\lambda\delta m_{max}\sqrt{\zeta} \left\| \mathbf{F}^j-\mathbf{F}^{k(j)}    \right\|))\\
=&\mathbb{E}((\mathbf{F}^j - \mathbf{F}^*)^2-vc(\left\|  \mathbf{F}^j - \mathbf{F}^{k(j)} \right\|^2 +\left\|  \mathbf{F}^{k(j)} - \mathbf{F}^* \right\|^2) \\
&+ 2v\delta\lambda m_{max} \sqrt{\zeta}(\left\|  \mathbf{F}^j - \mathbf{F}^{k(j)} \right\| +\left\|  \mathbf{F}^{k(j)} - \mathbf{F}^* \right\|)\\
&+2v^2M^2(3\rho\tau  + \frac{1}{2}))
\end{align*}

For the item $\mathbb{E}\left\|  \mathbf{F}^j - \mathbf{F}^{k(j)} \right\|^2 +\mathbb{E}\left\|  \mathbf{F}^{k(j)} - \mathbf{F}^* \right\|^2$, we can gain that

\begin{align*}
 &\mathbb{E}\left\|  \mathbf{F}^j - \mathbf{F}^{k(j)} \right\|^2 +\mathbb{E}\left\|  \mathbf{F}^{k(j)} - \mathbf{F}^* \right\|^2\\
 =&\mathbb{E}(\left\|  \mathbf{F}^j - \mathbf{F}^{*} \right\|^2  - ( \mathbf{F}^j - \mathbf{F}^{k(j)} )^T (\mathbf{F}^{k(j)} - \mathbf{F}^* ))\\
 \ge &\mathbb{E}(\left\|  \mathbf{F}^j - \mathbf{F}^{*} \right\|^2 -\sum_{i=k(j)}^{j-1} (\mathbf{F}^{i+1} - \mathbf{F}^{i})^T( \mathbf{F}^{k(j)} - \mathbf{F}^* ))\\
 \ge &\mathbb{E}(\left\|  \mathbf{F}^j - \mathbf{F}^{*} \right\|^2 -v\sum_{i=k(j)}^{j-1} \left\| A_{k(i)}P_{k(i)}L'(\mathbf{F}^{k(i)}) \right\| \left\| \mathbf{F}^{k(j)} - \mathbf{F}^*  \right\|)\\
 \ge & \mathbb{E}(\left\|  \mathbf{F}^j - \mathbf{F}^{*} \right\|^2 -v\sum_{i=k(j)}^{j-1}    M \left\| \mathbf{F}^{k(j)} - \mathbf{F}^*  \right\|)\\
  = & \mathbb{E}(\left\|  \mathbf{F}^j - \mathbf{F}^{*} \right\|^2 -v\tau   M \left\| \mathbf{F}^{k(j)} - \mathbf{F}^*  \right\|)
\end{align*}

Thus,
\begin{align*}
&\mathbb{E}(\mathbf{F}^{j+1}-\mathbf{F}^*)^2\\
\le&(1-vc)(\mathbb{E}(\mathbf{F}^j - \mathbf{F}^*))^2+ 2v\delta\lambda m_{max} \sqrt{\zeta} (\mathbb{E}\left\|  \mathbf{F}^j - \mathbf{F}^{k(j)} \right\|)\\ 
&+ (2v\delta\lambda m_{max}\sqrt{\zeta} +  v^2 c\tau M)(\mathbb{E}\left\|  \mathbf{F}^{k(j)} - \mathbf{F}^* \right\|)\\
&+2v^2M^2(3\rho\tau  + \frac{1}{2})\\
\le&(1-vc)\mathbb{E}(\mathbf{F}^j - \mathbf{F}^*)^2+ (4v\delta\lambda m_{max}\sqrt{\zeta} + cv^2\tau M)v\tau M\\ 
&+ (2v\delta\lambda m_{max} \sqrt{\zeta} + cv^2\tau M)(\mathbb{E}\left\|  \mathbf{F}^{j} - \mathbf{F}^* \right\|)\\
&+2v^2M^2(3\rho\tau  + \frac{1}{2})\\
=&(1-vc)\mathbb{E}(\mathbf{F}^j - \mathbf{F}^*)^2+  (2v\delta\lambda m_{max} \sqrt{\zeta} + cv^2\tau M)(\mathbb{E}\left\|  \mathbf{F}^{j} - \mathbf{F}^* \right\|)\\
&+(4v\delta\lambda m_{max}\sqrt{\zeta} + cv^2\tau M)v\tau M + 2v^2M^2(3\rho\tau  + \frac{1}{2})
\end{align*}

We define $C_1$, $C_2$ as follow
\begin{align*}
C_1 =  (2\delta\lambda m_{max} \sqrt{\zeta} + cv \tau M)\\
C_2 =  (4\delta\lambda m_{max}\sqrt{\zeta} + cv\tau M)\tau M + 2M^2(3\rho\tau  + \frac{1}{2})\\
\end{align*}


Above in-equation is quadratic recurrence for $\mathbb{E}\left\|  \mathbf{F}^{j} - \mathbf{F}^* \right\|^2$ and its convergence point is the fixed point.

In another word, this recurrence will convergent to $\mathbb{E}\left\| F_{\infty}-F^* \right\|^2$ in the contraction map rate $r$, where
\begin{align}
&\mathbb{E}\left\|  \mathbf{F}^{j+1} - \mathbf{F}^* \right\|^2 -\mathbb{E} \left\|  \mathbf{F}^{\infty} - \mathbf{F}^* \right\|^2\\& = r \left( \mathbb{E}\left\|  \mathbf{F}^{j} - \mathbf{F}^* \right\|^2 - \mathbb{E}\left\|  \mathbf{F}^{\infty} - \mathbf{F}^* \right\|^2  \right)\notag\\
\label{fixed}&\mathbb{E} \left\|  \mathbf{F}^{\infty} - \mathbf{F}^* \right\|^2=(\frac{C_1+\sqrt{C_1^2+4cvC_2}}{2c})^2 \\
\label{convergence_speed}&r=1-vc(1-\frac{C_1}{C_1+\sqrt{C_1^2+4cvC_2}}) 
\end{align}
 
To make our analysis clearly, we will make Eq.\ref{convergence_speed} into following format and only care about the term structure.

\begin{align}
r=1-vc(1-\frac{1}{1+\sqrt{1+C_3(C_4+\frac{C_5}{(\tau+C_6)+\frac{C_7}{\tau+C_6}+C_8})}}) \label{reformate_1}
\end{align} 

Where $C_3$ to $C_8$ are positive and do not contain $\tau$. In this form, we would know following result: when $\tau$ is large enough, with the increase of $\tau$, the convergence speed would decrease, i.e. $r$ would be increase. 

To gain the $r$'s sensitivity to $\tau$, the standard method is to treat $r$ as the function of $M,\tau,\rho$ and gain the $\nabla_\tau r(M,\rho,\tau)$. We expect that $\nabla_\tau r(M,\rho,\tau)$ is monotone to $M$ and $\rho$. However, the form of $\nabla_\tau r(M,\rho,\tau)$ is complex and it is exhausting to gain $\nabla_\tau r(M,\rho,\tau)$.

However, we notice that the coefficients of $\tau$ is the term about $M$, i.e. about $\Omega$ and $\rho$. Above factor shows that when   sampling rate is small, which leads to $M$ and $\rho$ are small, the influence of the changes of $\tau$ are small.

Thus, we also gain the result that small sampling rate, which leads to $M$ and $\rho$ are small, would reduce the convergence speed. This case is can be shown in Figure \ref{6vsverysamll}.

Because the coefficients of $\tau$ is also about $\zeta$, we can also conduct the conclusion that the setting whose  number of leaves entry is large would also reduce the influence of the changes of $\tau$. 

In Eq. \ref{fixed}, we can gain the result that Asynch-SGDBT would convergent to a range whose diameter is decided by the number of the tree, which decides $\delta$. A large range usually leads to a better generalization effect. 
 
When using the high instance diversity dataset and the GBDT setting GBDT trees contain massive leaves,  sampling operation leads to small $\rho$ and $M$, thus, high instance diversity dataset is apt to be accelerated. 
 
\subsection{conclusion}
 
 Based on the result from Eq. \ref{fixed} and Eq. \ref{convergence_speed}, we can easily gain following result:
 
 1. Asynch-SGBDT would converge to a range whose diameter is decided by the number of the tree. Usually We can gain a good generalization effect by using large diameter range.
 
 2.Small sampling rate leads to slow convergence speed. 
 
 3.Small sampling rate increase the scalability.
 
 4.Small learning rate increase the scalability.
 
 5.Small learning rate leads to slow convergence speed.  
 
 6. Asynch-SGBDT is apt to accelerate the GBDT  on a high-dimensional sparse dataset. 
 
 7. Asynch-SGBDT is apt to accelerate a GBDT whose trees contain massive leaves. 
 
\end{document}